\documentclass[10pt,twocolumn]{article}

\usepackage[letterpaper,margin=1in]{geometry}
\usepackage{times}
\usepackage{helvet}
\usepackage{courier}

\usepackage{cite}
\usepackage{amsmath,amssymb}
\usepackage{graphicx}
\usepackage{booktabs}
\usepackage{siunitx}
\usepackage{xurl}
\usepackage{microtype}
\usepackage{xcolor}
\usepackage{array}
\usepackage{algorithm}
\usepackage{algpseudocode}
\usepackage{placeins} 

\title{Constraint-Aware Discrete-Time PID Gain Optimization for Robotic Joint Control Under Actuator Saturation}

\renewcommand{\thefootnote}{\fnsymbol{footnote}}

\author{
Ojasva~Mishra$^{1}$
\quad Xiaolong~Wu$^{2}$
\quad Min~Xu$^{2}$\thanks{Corresponding author.}\\
$^{1}$Downingtown STEM Academy, Downingtown, PA, USA\\
$^{2}$Carnegie Mellon University, School of Computer Science, Pittsburgh, PA, USA
}

\begin{document}
\maketitle

\setcounter{footnote}{0}
\renewcommand{\thefootnote}{\arabic{footnote}}

\begin{abstract}
The precise regulation of rotary actuation is fundamental in autonomous robotics, yet practical PID loops deviate from continuous-time theory due to discrete-time execution, actuator saturation, and small delays and measurement imperfections. We present an implementation-aware analysis and tuning workflow for saturated discrete-time joint control. We (i) derive PI stability regions under Euler and exact zero-order-hold (ZOH) discretizations using the Jury criterion, (ii) evaluate a discrete back-calculation anti-windup realization under saturation-dominant regimes, and (iii) propose a hybrid-certified Bayesian optimization workflow that screens analytically unstable candidates and behaviorally unsafe transients while optimizing a robust IAE objective with soft penalties on overshoot and saturation duty. Baseline sweeps ($\tau=1.0$~s, $\Delta t=0.01$~s, $u\in[-10,10]$) quantify rise/settle trends for P/PI/PID. Under a randomized model family emulating uncertainty, delay, noise, quantization, and tighter saturation, robustness-oriented tuning improves median IAE from $0.687$ to $0.470$ while keeping median overshoot below $2\%$. In simulation-only tuning, the certification screen rejects $11.6\%$ of randomly sampled gains within bounds before full robust evaluation, improving sample efficiency without hardware experiments.
\end{abstract}

\vspace{0.35em}
\vspace{0.35em}

\section{Introduction}
The precise regulation of rotary actuation constitutes a fundamental
challenge in the engineering of autonomous robotic systems, ranging from
industrial manipulators to self-guided electric vehicle charging
mechanisms. In these cyber-physical domains, the fidelity of motion
control determines the operational success of the system. The
Proportional-Integral-Derivative (PID) controller remains a widely used standard in feedback control systems due to its
versatile structure and proven capacity to ensure asymptotic stability.

Despite the ubiquity of PID control, the translation of continuous-time
control theory into physical application is frequently complicated by
hardware constraints. Theoretical models often assume infinite control
authority and continuous signal processing; however, practical robotic
joints operate within strict voltage/torque limits and utilize digital
processors that execute control laws in discrete time steps. When a
controller demands an input that exceeds the physical capabilities of
the actuator, the system enters a regime of saturation, introducing
non-linearities that can degrade performance or induce instability.
In addition, real embedded loops exhibit non-idealities such as
computation delay, sensor noise, and encoder quantization that interact
with discretization and saturation.

This research presents a systematic characterization of a first-order
Linear Time-Invariant (LTI) robotic joint under discrete PID control.
Unlike generalized derivations found in introductory
literature, this study isolates the specific impact of control gains
(\(K_{p},\ K_{i},\ \text{and}\ K_{d}\)) within the context of a
magnitude-constrained system and then extends the analysis toward robust,
implementation-aware tuning in simulation.

\subsection*{Contributions}
While the baseline gain sweeps are intended to be interpretable, the paper is strengthened by three contributions that directly target robotics implementation gaps:
\begin{enumerate}
\item \textbf{Closed-form PI stability regions in discrete time.} For a first-order joint model, we derive explicit gain- and sampling-dependent stability constraints for PI control using the Jury criterion under both forward-Euler and exact ZOH discretizations, enabling ``safe'' gain selection before simulation.
\item \textbf{Saturation-dominant evaluation with anti-windup.} We include a discrete back-calculation anti-windup update and demonstrate how it changes recovery dynamics when saturation is frequent (a regime not captured by small-step, low-demand baselines).
\item \textbf{Robust constraint-aware gain selection.} We propose a robustness-oriented Bayesian gain-selection workflow that optimizes IAE while enforcing stability and soft constraints on overshoot and saturation duty across a randomized family of joint models (delay, noise, quantization, and parameter uncertainty).
\end{enumerate}

\section{Motivation}
\subsection{Problem formulation}
{\sloppy
While the Proportional-Integral-Derivative (PID) controller is the
theoretic standard for asymptotic stability in linear time-invariant
(LTI) systems, its implementation in robotic actuation is frequently
compromised by physical nonlinearities---specifically actuator
saturation and discretization artifacts. Theoretical derivations often
assume continuous-time signal processing and infinite control authority.
However, practical autonomous joints operate within strict voltage
limits ($u_{\min},u_{\max}$) and utilize digital processors that
execute control laws in discrete time steps (\(\Delta t\)). The
discrepancy between continuous-time theory and discrete-time
implementation creates a performance gap. When a high-gain controller
demands an input exceeding the actuator's physical capacity
($|u(t)| > u_{\mathrm{sat}}$), the system enters a regime of
saturation. This introduces windup phenomena and nonlinear transient
behaviors that degrade tracking fidelity, represented by the Integral of
Absolute Error (IAE) and settling time. Furthermore, the use of
first-order forward Euler discretization in low-latency embedded loops
introduces numerical approximations that deviate from ideal analytical
predictions \cite{fadali2020digital,zhang2023sampleddata}.
\par}

\subsection{Research objectives}
The primary objective of this study is to systematically characterize
the dynamic response of a first-order LTI robotic joint under
magnitude-constrained discrete PID control and to extend the analysis
toward robust practice without requiring hardware access.

Specifically, this research aims to:
\begin{enumerate}
\item Quantify discretization sensitivity: analyze the deviation between
 theoretical continuous-time predictions and discrete-time
 implementations using Forward Euler integration with limited sampling
 resolution (\(\Delta t = 0.01\,\text{s}\)).
\item Characterize saturation nonlinearities: evaluate the impact of control
 effort clamping (\(u_{k} \in [-10,10]\)) on integral
 windup and rise-time performance during step responses, including saturation-dominant tasks.
\item Derive constraint-aware tuning heuristics: establish tuning guidelines for \(K_{p}, K_{i}, K_{d}\)
 that minimize IAE while enforcing stability and bounded overshoot under model variations (delay, noise, parameter uncertainty).
\end{enumerate}

\begin{table}[!t]
\caption{Key symbols used throughout this paper with the equation number of introduction marked.}
\label{tab:symbols}
\centering
\footnotesize
\setlength{\tabcolsep}{5pt}
\renewcommand{\arraystretch}{1.06}
\begin{tabular}{@{}p{0.30\columnwidth}p{0.66\columnwidth}@{}}
\toprule
\textbf{Symbol} & \textbf{Description} \\
\midrule
$y(t),\,y_k$ & Joint output (position) in continuous and discrete time; plant output in \eqref{eq:plant}--\eqref{eq:euler}. \\
$u(t),\,u_k$ & Control effort; proportional/PI/PID laws in \eqref{eq:p}--\eqref{eq:pid}. \\
$r$ & Reference (desired joint position), unit step used throughout baseline experiments. \\
$e_k$ & Tracking error, $e_k=r-y_k$, defined after \eqref{eq:euler}. \\
$\tau$ & Plant time constant in the first-order model \eqref{eq:plant}. \\
$K$ & Plant static gain in \eqref{eq:plant}. \\
$\Delta t$ & Sampling interval (control period) in \eqref{eq:euler}. \\
$K_p,\,K_i,\,K_d$ & Proportional, integral, and derivative gains in \eqref{eq:p}--\eqref{eq:pid}. \\
$I_k$ & Discrete integral state and back-calculation update in \eqref{eq:antiwindup}. \\
$u_{\min},\,u_{\max}$ & Actuator saturation bounds used in the clamp \eqref{eq:sat}. \\
$\%OS$ & Percent overshoot metric in \eqref{eq:os}. \\
$t_r,\,t_s$ & Rise time and settling time in \eqref{eq:tr} and \eqref{eq:ts}. \\
$e_{ss}$ & Terminal steady-state error estimate in \eqref{eq:ess}. \\
$\mathrm{IAE}$ & Integral of absolute error in \eqref{eq:iae}. \\
$z$ & Discrete-time closed-loop pole for P control in \eqref{eq:cl_pole}. \\
$a,\,b$ & Exact ZOH discrete-time coefficients in \eqref{eq:zoh}. \\
$K_{aw}$ & Anti-windup (back-calculation) gain in \eqref{eq:antiwindup}. \\
$J$ & Robust BO objective in \eqref{eq:bo_obj_robust}. \\
$\lambda_{os},\,\lambda_{sat},\,\lambda_u$ & BO penalty weights in \eqref{eq:bo_obj_robust}. \\
\bottomrule
\end{tabular}
\end{table}

\section{Related work}
This paper connects three threads that frequently appear in robotics actuation but are often studied in isolation: (i) digital PID implementation effects (sampling and discretization), (ii) saturation and anti-windup behavior in voltage/torque-limited joints, and (iii) data-efficient gain selection under uncertainty when direct hardware iteration is impractical.

\subsection{Digital PID in robotics and sampled-data effects}
PID remains a workhorse in robotics because it is simple, interpretable, and effective across a wide range of actuation plants. Recent reviews and sampled-data surveys emphasize that real closed-loop behavior depends strongly on implementation details that are frequently abstracted away in continuous-time analysis, including sampling, numerical discretization, delays, quantization, and actuator limits \cite{borase2021pid,zhang2023sampleddata}. In embedded joint control, the sampling period is not a nuisance parameter: it directly shapes phase margin and can turn a seemingly benign continuous-time tuning into an unstable discrete-time loop.

Most classical PID tuning recipes are formulated in continuous time and then discretized heuristically. In contrast, we explicitly characterize discrete-time stability for the PI subset under forward-Euler discretization, yielding gain- and sampling-dependent stability regions (Sec.~\ref{sec:pi_stability}). This complements simulation sweeps by providing an analytic constraint that can be checked before running expensive trials and that can be used to prune unsafe regions of the search space.

\subsection{Saturation, integral windup, and anti-windup compensation}
When a commanded effort exceeds actuator bounds, the loop becomes nonlinear and integral windup can dominate settling behavior, producing long recovery tails and overshoot once the actuator leaves saturation. Contemporary anti-windup work continues to refine compensation mechanisms (including back-calculation and two-stage designs) that explicitly account for saturation, quantization, and embedded implementation constraints \cite{leal2020doublebackcalc,borja2021saturation,turner2024aw}. In robotics, where large setpoint steps and load disturbances are common, windup is not a rare failure mode but a routine operating condition.

Our contribution is not a new anti-windup law, but a robotics-oriented evaluation: we isolate a saturation-dominant regime and quantify how standard back-calculation reshapes the transient, including recovery time and overshoot (Sec.~\ref{sec:aw_results}). This provides a grounded baseline for anti-windup-aware gain optimization, where the objective must reflect saturation behavior rather than nominal linear dynamics.

\subsection{Data-driven and safe controller tuning}
Derivative-free tuning has grown rapidly because it can optimize task-specific objectives even when the plant is hard to model precisely. Bayesian optimization (BO) is attractive for controller tuning because it is sample-efficient and supports noisy, nonconvex objectives via probabilistic surrogate models \cite{garnett2023bo,wang2023advbo,balandat2020botorch,fujimoto2023review}. Recent control-engineering work demonstrates BO for systematic PID tuning, including multi-loop digital PID design and careful construction of bounded search domains \cite{coutinho2023pidbo,zagorowska2025bo,khosravi2020cascade}. 

For robotics, the key challenge is not only efficiency but also \emph{safety}: many candidate gains lead to instability, violent overshoot, or sustained saturation. Safety-aware BO and constrained BO provide mechanisms for restricting evaluations to feasible regions, including probabilistic safety constraints and time-varying formulations \cite{eriksson2021scbo,berkenkamp2021safe,brunzema2022tvbo,koenig2023ragoose,vonrohr2024crash,widmer2023legged_sbo}. Our hybrid certification (Sec.~\ref{sec:bo_workflow}) follows this line of work but is intentionally lightweight: it couples analytic PI stability checks with a runtime divergence detector so that unsafe trials are terminated quickly rather than corrupting the BO dataset.

\subsection{Simulation benchmarks as hardware surrogates}
Because access to physical hardware is often limited, recent robotics research increasingly uses benchmark suites and high-throughput simulation to evaluate robustness under uncertainty and to compare safe tuning methods \cite{safecontrolgym2022,mower2022ropi,makoviychuk2021isaacgym}. Our robustness ensemble and expanded benchmark suite adopt the same philosophy: rather than relying on a single nominal plant, we evaluate gains across uncertainty in dynamics, delays, sensing noise, and actuator limits, and we report both best-case performance and outlier behavior relevant to deployment.

\section{Design and methodology}
\label{sec:design}
This section specifies the plant and actuator models, discrete-time implementation, saturation handling, and evaluation metrics used throughout the paper. It then derives discrete-time PI stability regions via the Jury criterion under both Euler and exact ZOH discretizations and formalizes the constrained objective used for robust gain selection. These components define the feasibility and scoring framework used in the baseline sweeps, saturation-dominant experiments, Monte Carlo robustness evaluation, and constrained Bayesian optimization results.

\subsection{Modeling and discrete-time implementation}
\label{sec:modeling}

\subsubsection{Variables and conventions}
The input $u(t)$ is the control effort applied to the robotic joint. The output $y(t)$ is the measured joint position. The plant has static gain $K>0$ and time constant $\tau>0$. The sampling interval is $\Delta t$ with sample index $k$ and sample times $t_k=k\Delta t$. The tracking error at sample $k$ is $e_k=r-y_k$. The controller gains are $K_p$ (proportional), $K_i$ (integral), and $K_d$ (derivative).

Unless stated otherwise, constants are $\tau=1.0$ s, $K=1.0$, $\Delta t=0.01$ s, total time $T=5.0$ s, reference $r=1$, and clamp $u_k\in[-10,10]$.

\subsubsection{Plant model and continuous-time step response}
A first-order LTI model approximates actuator dynamics:
\begin{equation}
\tau\frac{dy(t)}{dt}+y(t)=Ku(t).
\label{eq:plant}
\end{equation}
For a unit-step input $u(t)=1$ and $y(0)=0$, the step response is
\begin{equation}
y(t)=K\left(1-e^{-t/\tau}\right).
\label{eq:plant_step}
\end{equation}

\subsubsection{Forward-Euler discretization}
Robotic controllers execute in discrete time. Using a forward-Euler update, the plant becomes:
\begin{equation}
y_{k+1}=y_k+\Delta t\left(-\frac{1}{\tau}y_k+\frac{K}{\tau}u_k\right).
\label{eq:euler}
\end{equation}
The tracking error is $e_k=r-y_k$.

\subsubsection{Controller laws}
\textbf{P control:}
\begin{equation}
u_k=K_p e_k.
\label{eq:p}
\end{equation}
\textbf{PI control:}
\begin{equation}
u_k=K_p e_k + K_i I_k,\quad I_{k+1}=I_k+e_k\Delta t.
\label{eq:pi}
\end{equation}
\textbf{PID control:}
\begin{equation}
u_k=K_p e_k + K_i I_k + K_d\frac{e_k-e_{k-1}}{\Delta t}.
\label{eq:pid}
\end{equation}

\subsubsection{Actuator saturation, derivative filtering, and discrete anti-windup}
In hardware, the actuator saturates:
\begin{equation}
u_k^{\mathrm{sat}}=\mathrm{clip}\big(u_k, u_{\min}, u_{\max}\big).
\label{eq:sat}
\end{equation}
The derivative term is well known to amplify measurement noise; practical PID implementations therefore use a filtered derivative (band-limited differentiator) rather than an ideal discrete difference \cite{borase2021pid,zhang2023sampleddata}. In our pipeline, the derivative estimate is passed through a first-order low-pass filter; this does not change the conceptual derivations, but it yields behavior closer to embedded practice in the delay/noise experiments.

To explicitly model embedded control under saturation, we include a discrete back-calculation anti-windup augmentation:
\begin{equation}
\begin{aligned}
I_{k+1} &= I_k+\Delta t\left(e_k + \frac{u_k^{\mathrm{sat}}-u_k}{K_{aw}}\right),\\
u_k &= K_p e_k+K_i I_k + K_d\frac{e_k-e_{k-1}}{\Delta t},
\end{aligned}
\label{eq:antiwindup}
\end{equation}
where $K_{aw}>0$ parameterizes an \emph{inverse} correction gain (equivalently, a time constant $T_{aw}=K_{aw}$); larger $K_{aw}$ weakens the correction and $K_{aw}\rightarrow\infty$ recovers the baseline integrator.

\subsection{Performance metrics}
\label{sec:metrics}
To compare controllers across gain sweeps and across non-ideal model variants, we report a small set of standard step-response and tracking-fidelity metrics. All metrics are computed from sampled trajectories $\{y_k\}_{k=0}^{N}$ at times $t_k=k\Delta t$ with reference $r$.

\textbf{Percent overshoot (\%OS).}
Overshoot quantifies the maximum peak above the reference during a transient. It is reported as a percentage of $r$ and clipped at zero so monotone responses do not produce negative overshoot:
\begin{align}
\%OS &= \max\!\left(0,\frac{\max_k(y_k)-r}{r}\right)\cdot 100. \label{eq:os}
\end{align}
For the delay-free first-order plant, overshoot is typically zero in the stable gain range; overshoot becomes meaningful once delay, second-order dynamics, or saturation-driven windup are present.

\textbf{Rise time ($t_r$).}
Rise time measures how quickly the output approaches the target. We use the 90\% threshold definition:
\begin{align}
t_r &= \min\{t_k:\ y_k\ge 0.9r\}. \label{eq:tr}
\end{align}
If the trajectory does not reach $0.9r$ within the finite simulation horizon, we report $t_r$ as ``--'' in tables.

\textbf{Settling time ($t_s$).}
Settling time measures when the trajectory enters and stays within a tight tolerance band around $r$. We use a strict $\pm 2\%$ band:
\begin{align}
t_s &= \min\left\{t_k:\ |y_j-r|\le 0.02r\ \forall j\ge k\right\}. \label{eq:ts}
\end{align}
If the response never satisfies the band condition over the horizon, we report ``--'' (and in some plots we annotate as $>T$).

\textbf{Terminal steady-state error ($e_{ss}$).}
Because experiments run for a finite horizon, we estimate terminal tracking error by averaging the absolute error over the last $N_{\mathrm{end}}$ samples:
\begin{align}
e_{ss} &\approx \frac{1}{N_{\mathrm{end}}}\sum_{k=N-N_{\mathrm{end}}+1}^{N} |r-y_k|. \label{eq:ess}
\end{align}

\textbf{Integral of absolute error (IAE).}
IAE measures cumulative tracking error over the full trajectory and directly penalizes long recovery tails:
\begin{align}
\mathrm{IAE} &\approx \sum_{k=0}^{N}|e_k|\Delta t. \label{eq:iae}
\end{align}

\subsection{Discrete-time stability regions for PI control}
\label{sec:pi_stability}

To connect the PI stability results to the familiar proportional case, consider P control with Euler discretization.
Combining \eqref{eq:euler} and \eqref{eq:p} (with $r=0$ for stability analysis) yields the scalar closed-loop pole
\begin{equation}
z = 1-\frac{\Delta t}{\tau}\left(1+K K_p\right),
\label{eq:cl_pole}
\end{equation}
and stability requires $|z|<1$, which implies $0<\Delta t<\frac{2\tau}{1+K K_p}$.

The baseline section uses simulations to reveal trends, but safe gain selection for embedded control benefits from \emph{closed-form} stability constraints. For PI control, the closed-loop sampled-data system is second-order, and the Jury criterion provides necessary and sufficient conditions for discrete-time stability \cite{fadali2020digital}.

\subsubsection{Euler discretization: explicit Jury inequalities}
\begin{figure}[!t]
\centering
\includegraphics[width=\columnwidth]{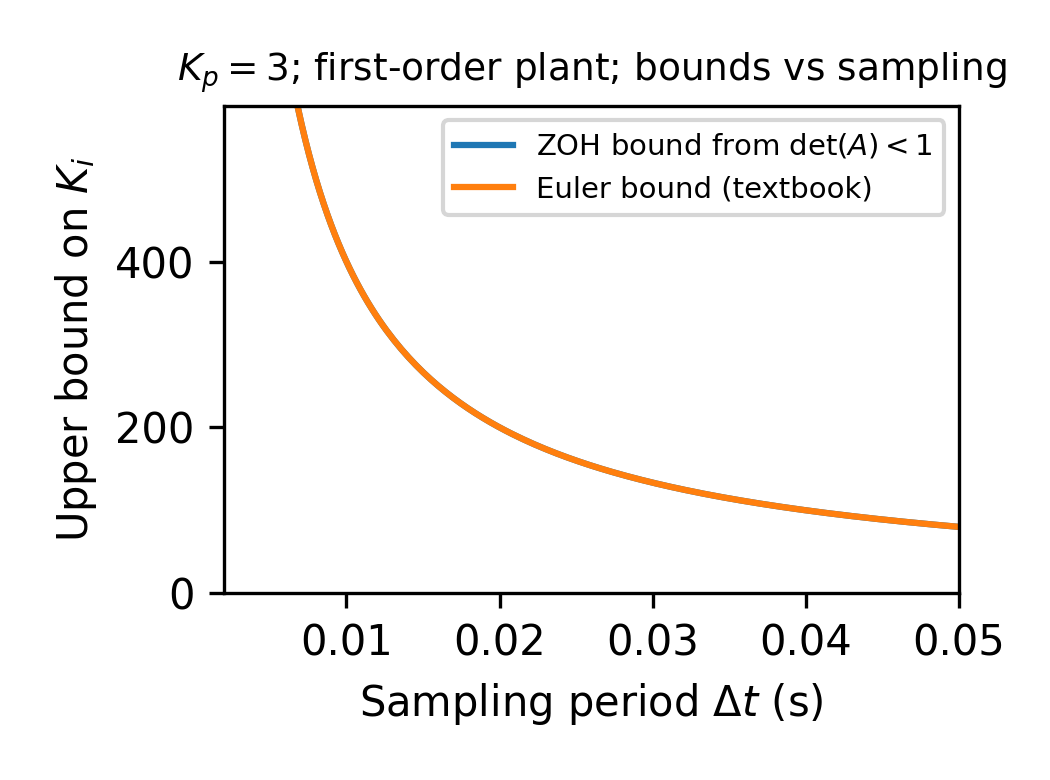}
\caption{Sampling-period dependence of the integrator gain guardrail for a first-order plant with PI control. The forward-Euler condition in \eqref{eq:ki_upper} is compared to a ZOH-derived sufficient bound obtained from the discrete-time determinant condition (Sec.~III-B).}
\label{fig:pi_stab_euler}
\end{figure}

Let $\alpha=\Delta t/\tau$. For stability analysis set $r=0$ so $e_k=-y_k$. Under Euler discretization, the closed-loop state update for $x_k=[y_k, I_k]^T$ becomes
\begin{equation}
x_{k+1}=
\begin{bmatrix}
1-\alpha(1+KK_p) & \alpha K K_i\\
-\Delta t & 1
\end{bmatrix}x_k.
\end{equation}
Let $\mathrm{tr}$ and $\det$ be the trace and determinant of the state matrix. The characteristic polynomial is $\lambda^2-\mathrm{tr}\,\lambda+\det=0$. Applying the Jury test for a monic second-order polynomial yields three inequalities:
\begin{align}
K_i &> 0, \label{eq:jury1}\\
4-2\alpha(1+KK_p)+\alpha K K_i \Delta t &> 0, \label{eq:jury2}\\
\alpha(1+KK_p) - \alpha K K_i \Delta t &> 0. \label{eq:jury3}
\end{align}
The final inequality can be rearranged into an interpretable guardrail:
\begin{equation}
K_i < \frac{1+KK_p}{K\,\Delta t}.
\label{eq:ki_upper}
\end{equation}
Fig.~\ref{fig:pi_stab_euler} highlights how the integrator-gain guardrail varies with sampling, and why discretization choice changes the admissible $K_i$ at practical $\Delta t$.

\subsubsection{Exact ZOH discretization}
\begin{figure}[!t]
\centering
\includegraphics[width=\columnwidth]{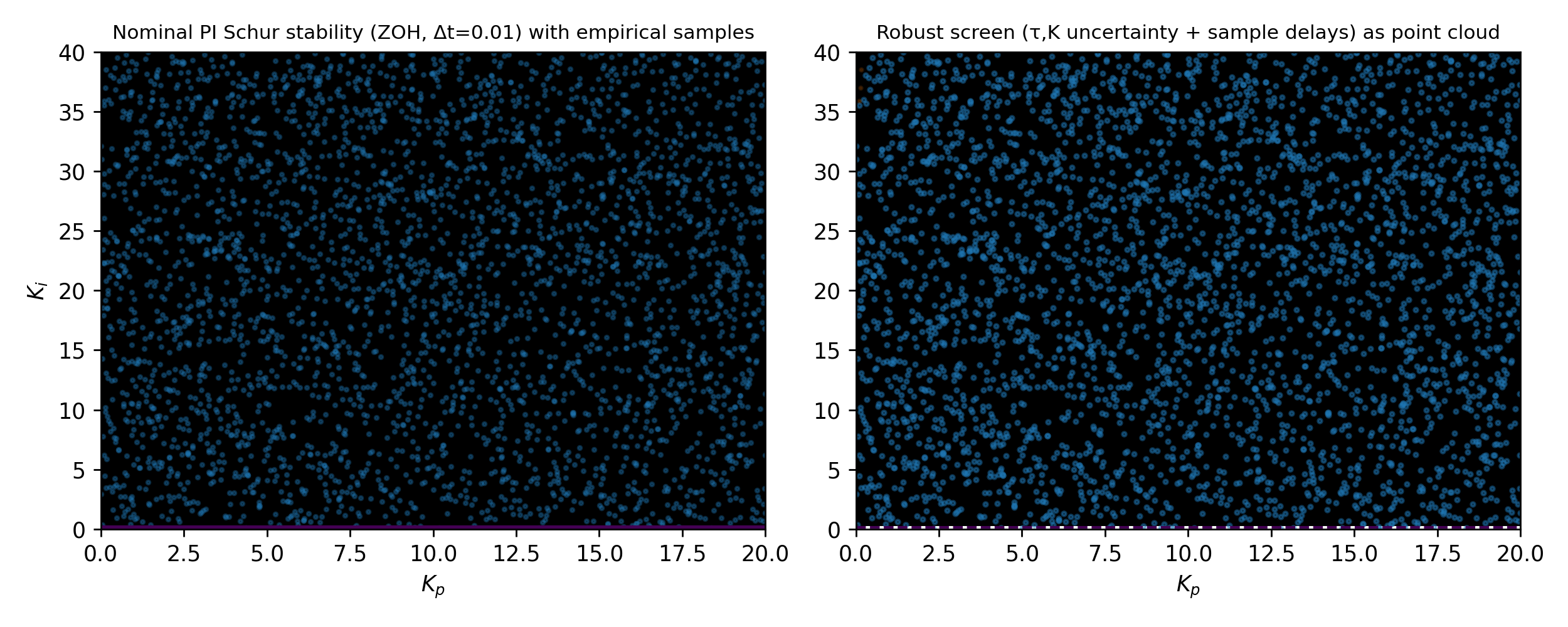}
\caption{Nominal PI Schur-stable set (ZOH, $\Delta t=0.01$) and robust point-cloud screen under plant uncertainty and sample delays; sampled gain outcomes are overlaid to validate the analytic boundary.}
\label{fig:pi_stab_zoh}
\end{figure}

Under a ZOH assumption, the exact discrete plant is $y_{k+1}=ay_k+bu_k$ with $a=e^{-\Delta t/\tau}$ and $b=K(1-a)$. The closed-loop PI state matrix becomes
\begin{equation}
x_{k+1}=
\begin{bmatrix}
a-bK_p & bK_i\\
-\Delta t & 1
\end{bmatrix}x_k,
\end{equation}
and the same Jury logic yields a comparable stability region (Fig.~\ref{fig:pi_stab_zoh}).

\subsection{Saturation-dominant behavior and anti-windup}
\label{sec:aw_design}
\label{sec:sat_dominant}
To evaluate saturation and windup directly (rather than only in the baseline sweeps), we define a saturation-dominant scenario by (i) tightening the actuator clamp so that early transients saturate, and (ii) adding a small input delay and lightly damped second-order actuator dynamics to emulate realistic servo behavior. In this regime, integral windup is expected to elongate recovery tails unless anti-windup is applied. The experiment reports both the output trajectory and the commanded/saturated effort to directly visualize clamping and recovery.

\subsection{Robust constraint-aware gain tuning}
\label{sec:robust_design}
To quantify robustness under non-idealities, we evaluate gains across a randomized family of models:
\begin{itemize}
\item Uncertain first-order parameters: $\tau\sim \mathcal{U}[0.5,1.5]$ and $K\sim \mathcal{U}[0.8,1.2]$.
\item Input delay: $d\in\{0,1,2,3\}$ samples.
\item Measurement imperfections: additive noise $\sigma\sim \mathcal{U}[0,0.01]$ and quantization $\Delta_q\in\{0,0.001,0.002\}$.
\item Tighter saturation limits: $u_{\max}\in\{2,3,5\}$ to trigger clamping in a subset of trials.
\end{itemize}
Each candidate gain triple is scored by an objective that aggregates both nominal tracking and robustness:
\begin{equation}
\begin{aligned}
J(K_p,K_i,K_d) &= \mathrm{median}_{m\in\mathcal{M}}\, J_m,\\
J_m &= \mathrm{IAE}_m
+ \lambda_{os}\,\max(0,\%OS_m-\%OS_{\max})^2 \\
&\quad + \lambda_{sat}\,\mathrm{sat\_duty}_m^2
+ \lambda_u\,u_{\mathrm{rms},m}^2.
\end{aligned}
\label{eq:bo_obj_robust}
\end{equation}
\noindent\textbf{Parameterization.} Unless noted, we set $\%\mathrm{OS}_{\max}=5\%$, $(\lambda_{os},\lambda_{sat},\lambda_{u})=(1,5,0.5)$, and compute all metrics over a $2$~s horizon with unit-step reference; $\mathrm{IAE}$ is normalized by the horizon length to keep $J$ dimensionless.
where $\mathcal{M}$ indexes randomized models and $\mathrm{sat\_duty}_m$ is the fraction of time the actuator saturates. This objective is compatible with Bayesian optimization (BO) because it is expensive, noisy, and non-convex.

\subsection{ML-aided gain tuning via Bayesian optimization}
\label{sec:bo_workflow}
Manual tuning requires iterative trial-and-error. An alternative is to treat the closed-loop evaluation as a black box and optimize gains by minimizing a scalar cost function built from \eqref{eq:os}--\eqref{eq:iae} (or the robust objective in \eqref{eq:bo_obj_robust}). Bayesian optimization iteratively proposes $(K_p,K_i,K_d)$, evaluates $J$, and updates a surrogate model to select the next candidate \cite{wang2023advbo,garnett2023bo}.

\begin{figure}[!t]
\centering
\includegraphics[width=\columnwidth]{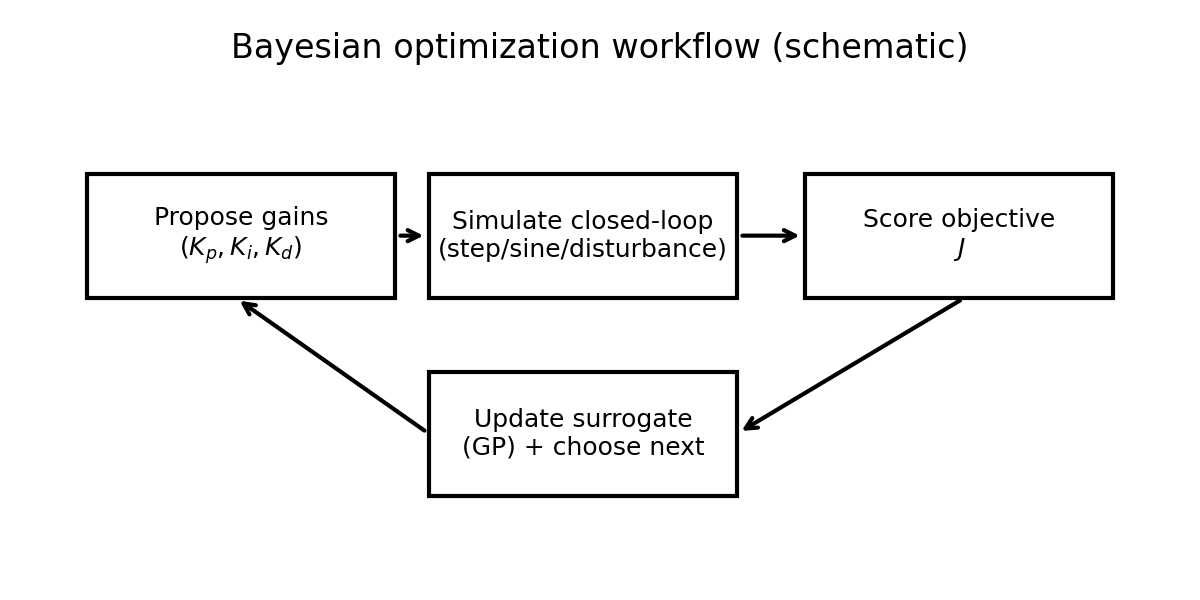}
\caption{BO workflow: propose gains, evaluate, score $J$, update surrogate.}
\label{fig:ml_schematic}
\end{figure}

\subsubsection{Hybrid certification for safe gain exploration}
A central risk of black-box gain tuning is evaluating controllers that are numerically stable under saturation yet dynamically unacceptable (e.g., extreme overshoot, sustained saturation, or long windup recovery). We introduce a hybrid certification filter that prevents such evaluations while retaining sample efficiency:
\begin{itemize}
\item \textbf{Analytic feasibility (stability):} we restrict candidates to the ZOH Jury-feasible PI region $\mathcal{S}$ derived in Sec.~\ref{sec:pi_stability}. This provides a conservative, interpretable guardrail for sampled-data stability of the nominal first-order joint model.
\item \textbf{Behavioral certification (constraints):} we additionally require that a short-horizon nominal check on a second-order actuator model satisfies bounded overshoot and does not remain saturated for the entire transient. This rejects aggressively destabilizing gains before expensive robust evaluation.
\end{itemize}
This filter is lightweight and evaluation-driven: it enforces robotics-relevant behavior constraints rather than merely preventing divergence.

\begin{algorithm}[!t]
\caption{Hybrid-certified Safe-BO (HC-SBO) for PID gains.}
\label{alg:hcsbo}
\begin{algorithmic}[1]
\State \textbf{Input:} gain bounds $\mathcal{B}$; analytic feasible set $\mathcal{S}$ (Jury); behavioral filter $\mathcal{F}$ (short-horizon check); robust objective $J(\cdot)$; initial budget $n_0$; total budget $N$.
\State Sample $n_0$ points from $\mathcal{B}$ until each satisfies $x\in(\mathcal{S}\cap\mathcal{F})$; evaluate $J(x)$.
\For{$n=n_0$ to $N-1$}
  \State Fit GP surrogate $\hat{J}$.
  \State Draw candidate pool $\mathcal{C}\subset \mathcal{B}$ and keep only feasible candidates $\mathcal{C}\leftarrow\{x\in\mathcal{C}: x\in\mathcal{S}\cap\mathcal{F}\}$.
  \State Choose $x_{n+1}=\arg\max_{x\in\mathcal{C}} \mathrm{EI}(x;\hat{J})$.
  \State Evaluate $J(x_{n+1})$ and update dataset.
\EndFor
\State \textbf{Output:} best gains observed.
\end{algorithmic}
\end{algorithm}

\noindent\textbf{PI guard vs. PID tuning.} The analytic Jury guard applies to the PI pair $(K_p,K_i)$ and is used as a conservative prefilter; the derivative term is safeguarded by the behavioral screen and the robust evaluation stage.

\subsubsection{Actuator-realistic second-order joint model}
\label{sec:second_order_bench}
The first-order plant in \eqref{eq:plant} captures a velocity-limited actuator, but many robotic joints exhibit second-order behavior dominated by inertia and damping. To emulate this, we evaluate gains on the normalized second-order actuator model
\begin{equation}
\ddot{\theta}(t) + 2\zeta\omega_n \dot{\theta}(t) + \omega_n^2 \theta(t) = K_u\,u(t) + d(t),
\label{eq:second_order}
\end{equation}
where $\theta$ is joint position, $(\omega_n,\zeta)$ set natural frequency and damping, $K_u$ maps input to torque/acceleration, and $d(t)$ is a disturbance torque. We augment \eqref{eq:second_order} with practical non-idealities: (i) input delay of 1--2 samples, (ii) Coulomb and viscous friction, (iii) a small deadzone, (iv) encoder quantization and additive noise, and (v) tight saturation that makes anti-windup essential. This model bridges a single-pole surrogate and actuator behavior commonly observed in robotics.

\subsubsection{Expanded robust benchmark suite}
We define a robust benchmark that combines (A) the ZOH-discretized first-order joint and (B) the second-order actuator model \eqref{eq:second_order}. For each family, parameters are sampled from bounded uncertainty sets and evaluated under three tasks: step tracking, sinusoidal tracking, and disturbance rejection. The reported objective is the median score over draws and tasks; it aggregates tracking error (IAE/RMSE), overshoot penalties, and control-effort penalties.

\section{Experiments and results}
\label{sec:experiments}
This section reports baseline gain sweeps and then progressively more realistic evaluation results (saturation-dominant tests, Monte Carlo robustness, and expanded-benchmark BO results).

\subsection{Evaluation validity and scope}
\label{sec:sim_validity}
The results are reported as comparative outcomes under a consistent evaluation protocol. All controllers are executed in discrete time with ZOH actuation and explicit effort clamping, and all candidate gains are tested against non-idealities that commonly dominate PID performance: sample delay, measurement noise, encoder quantization, and tighter saturation limits (Sec.~\ref{sec:robust_design}). In addition to a first-order surrogate, gains are evaluated on a lightly damped second-order actuator family with friction and deadzone (Sec.~\ref{sec:second_order_bench}). These evaluations quantify typical behavior and violation rates under uncertainty rather than relying on a single nominal trajectory.

\begin{table}[!t]
\caption{Modeled non-idealities used to emulate embedded joint control.}
\label{tab:sim_nonidealities}
\centering
\resizebox{\columnwidth}{!}{%
\begin{tabular}{@{}p{0.26\columnwidth}p{0.34\columnwidth}p{0.34\columnwidth}@{}}
\toprule
\textbf{Non-ideality} & \textbf{Why it matters} & \textbf{How modeled (this paper)} \\
\midrule
Effort saturation & current/voltage limits cause nonlinear recovery and windup & hard clamp $u\in[u_{\min},u_{\max}]$; saturation-dominant setting tightens bounds (Sec.~\ref{sec:sat_dominant}) \\
Sample delay & compute/communication delay reduces phase margin & $d\in\{0,1,2,3\}$ samples in robustness ensemble \\
Noise \& quantization & sensors/encoders inject jitter; $K_d$ can amplify it & additive noise $\sigma\sim\mathcal{U}[0,0.01]$ and quantization $\Delta_q\in\{0,0.001,0.002\}$ (Sec.~\ref{sec:robust_design}) \\
Second-order dynamics & motors/links add lightly damped modes & actuator family in \eqref{eq:second_order} with $(\omega_n,\zeta)$ and input gain $K_u$ (Sec.~\ref{sec:second_order_bench}) \\
Friction \& deadzone & stiction and Coulomb friction distort small motions & viscous $b=0.05$--$0.06$, Coulomb $f_c=0.02$--$0.03$, small deadzone (Table~\ref{tab:second_order_params}) \\
\bottomrule
\end{tabular}}
\end{table}

\subsection{Experimental design (baseline)}
\label{sec:baseline_design}
We perform three sweeps:
\begin{itemize}
\item P only: $K_p\in\{0.5,1.0,1.5,2.0,3.0\}$ with $K_i=0$, $K_d=0$.
\item PI: fix $K_p=3.0$ and vary $K_i\in\{0.00,0.25,0.50,1.00\}$ with $K_d=0$.
\item PID: fix $K_p=3.0$, $K_i=1.0$ and vary $K_d\in\{0.00,0.05,0.10\}$.
\end{itemize}
For each run we log $t_k$, $y_k$, $u_k$, and $e_k$, compute the metrics in \eqref{eq:os}--\eqref{eq:iae}, and produce plots and tables.

\subsection{Results (baseline characterization)}
\label{sec:baseline_results}

\subsubsection{Proportional-only control}

\begin{table}[!t]
\caption{P-only step-response metrics ($K_i=0$, $K_d=0$).}
\label{tab:p_only}
\centering
\resizebox{\columnwidth}{!}{%
\begin{tabular}{@{}rrrrrr@{}}
\toprule
$K_p$ & \%OS & $t_r$ (s) & $t_s$ (s) & $e_{ss}$ & IAE \\
\midrule
0.5 & 0.0000 & -- & -- & 0.6669 & 3.5495 \\
1.0 & 0.0000 & -- & -- & 0.5000 & 2.7419 \\
1.5 & 0.0000 & -- & -- & 0.4000 & 2.2306 \\
2.0 & 0.0000 & -- & -- & 0.3333 & 1.8787 \\
3.0 & 0.0000 & -- & -- & 0.2500 & 1.4263 \\
\bottomrule
\end{tabular}}
\end{table}

\begin{figure}[!t]
\centering
\includegraphics[width=\columnwidth]{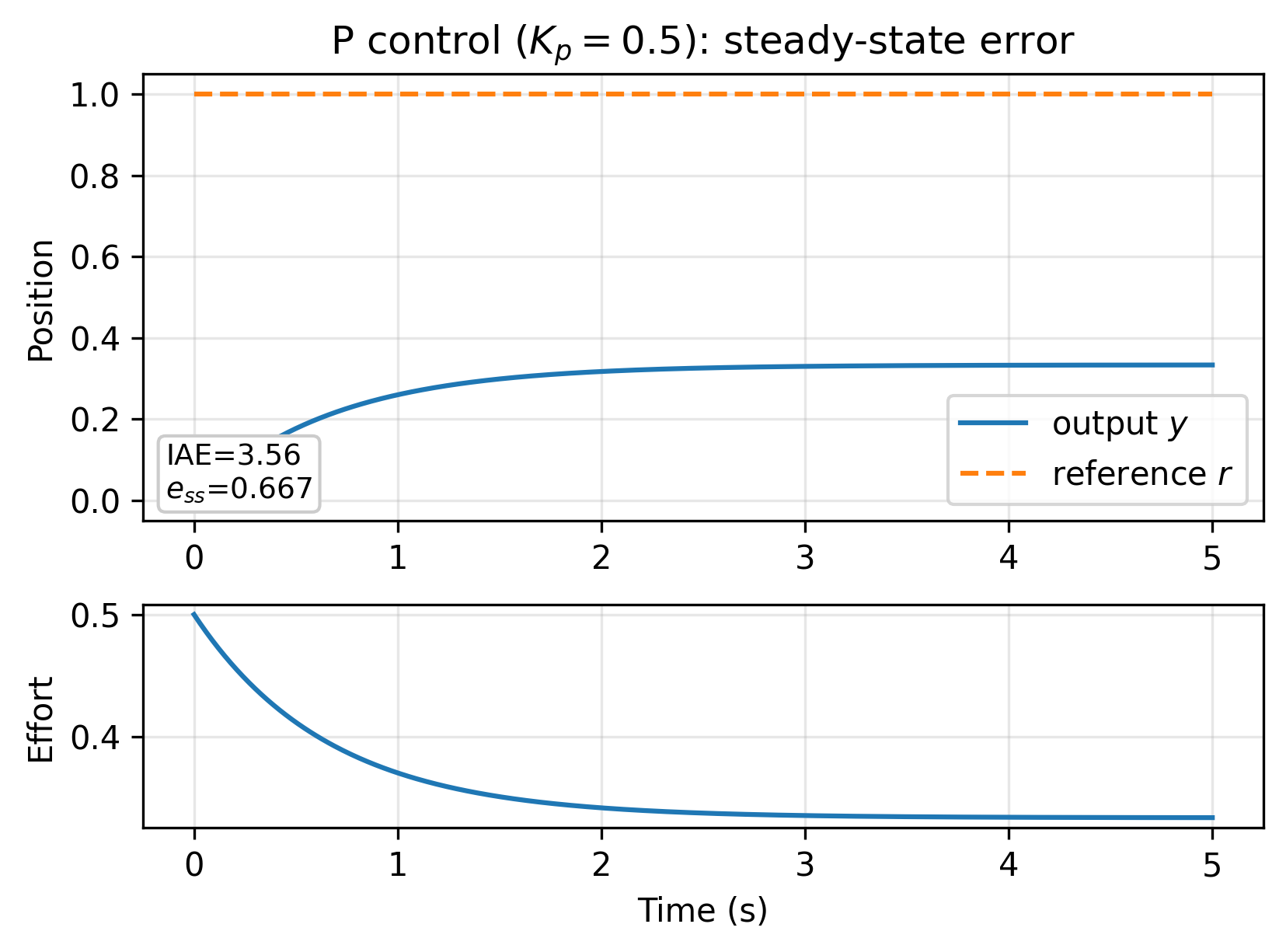}
\caption{P-only step response at $K_p=0.5$.}
\label{fig:p_kp05}
\end{figure}

\begin{figure}[!t]
\centering
\includegraphics[width=\columnwidth]{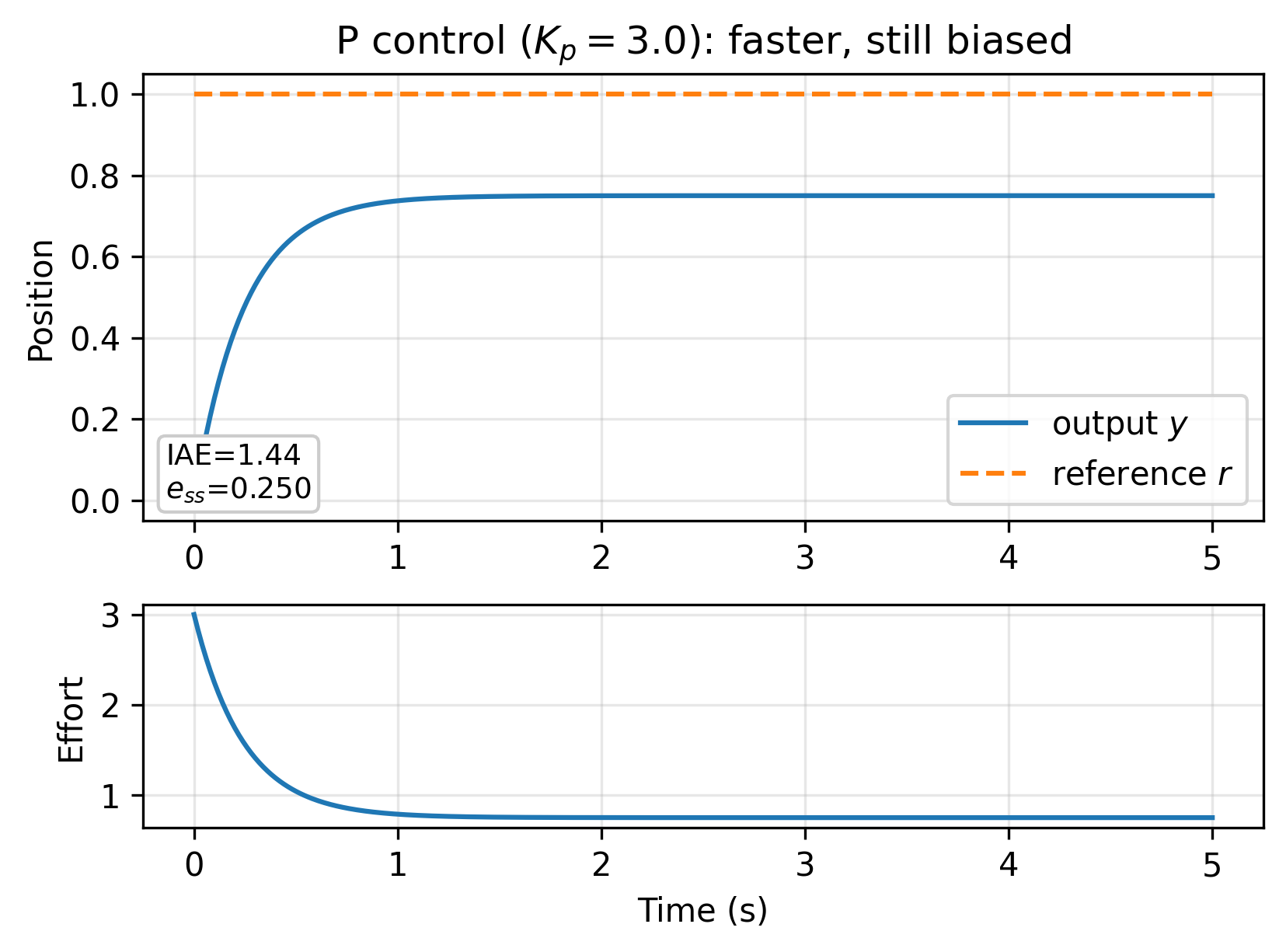}
\caption{P-only step response at $K_p=3.0$.}
\label{fig:p_kp30}
\end{figure}

Figs.~\ref{fig:p_kp05} and \ref{fig:p_kp30} illustrate monotone convergence and the expected reduction of the steady-state tail as $K_p$ increases. Overshoot is identically zero across this sweep for the delay-free single-pole plant.

For a stable first-order plant without delay, P control produces a monotone, non-oscillatory transient because the closed-loop dynamics remain first-order. The key limitation is the nonzero steady-state error: with $u=K_p(r-y)$ and plant DC gain $K$, the closed-loop DC gain is $\frac{KK_p}{1+KK_p}$, so for a unit step $r=1$ the asymptotic output is $y_\infty=\frac{KK_p}{1+KK_p}$ and the steady-state error is $e_\infty=\frac{1}{1+KK_p}$. This directly explains the trend in Table~\ref{tab:p_only}: as $K_p$ increases, $e_{ss}$ decreases approximately like $\frac{1}{1+K_p}$ for $K=1$, and IAE decreases because the tail area shrinks.

\subsubsection{Proportional--integral control}

\begin{table}[!t]
\caption{PI step-response metrics at $K_p=3.0$ ($K_d=0$).}
\label{tab:pi}
\centering
\resizebox{\columnwidth}{!}{%
\begin{tabular}{@{}rrrrrr@{}}
\toprule
$K_i$ & \%OS & $t_r$ (s) & $t_s$ (s) & $e_{ss}$ & IAE \\
\midrule
0.00 & 0.0000 & -- & -- & 0.2500 & 1.4263 \\
0.25 & 0.0000 & -- & -- & 0.1786 & 1.2159 \\
0.50 & 0.0000 & -- & -- & 0.1257 & 1.0419 \\
1.00 & 0.0000 & 2.7800 & -- & 0.0590 & 0.7789 \\
\bottomrule
\end{tabular}}
\end{table}

\begin{figure}[!t]
\centering
\includegraphics[width=\columnwidth]{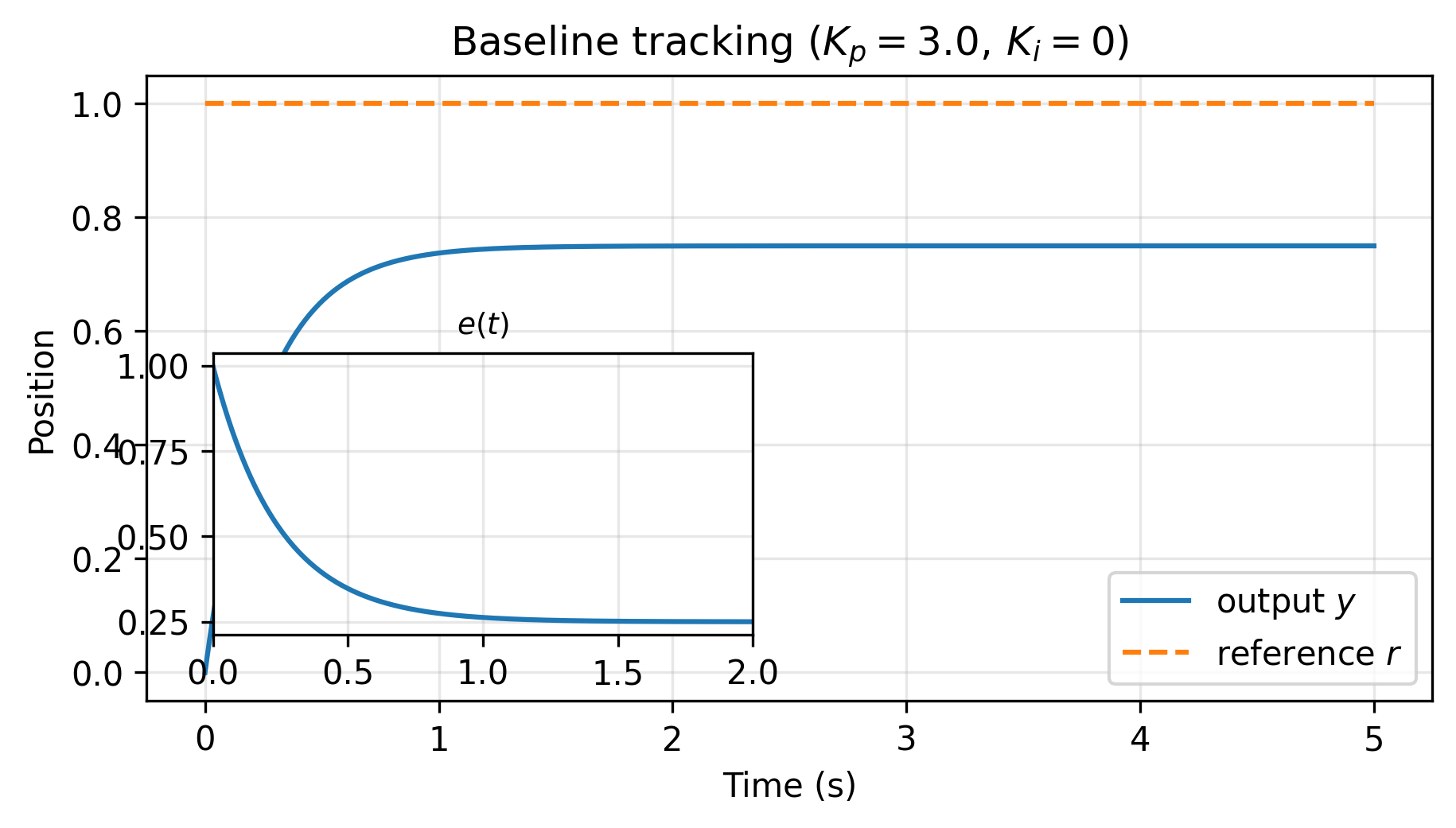}
\caption{Baseline ($K_p=3.0$, $K_i=0$).}
\label{fig:pi_baseline}
\end{figure}

\begin{figure}[!t]
\centering
\includegraphics[width=\columnwidth]{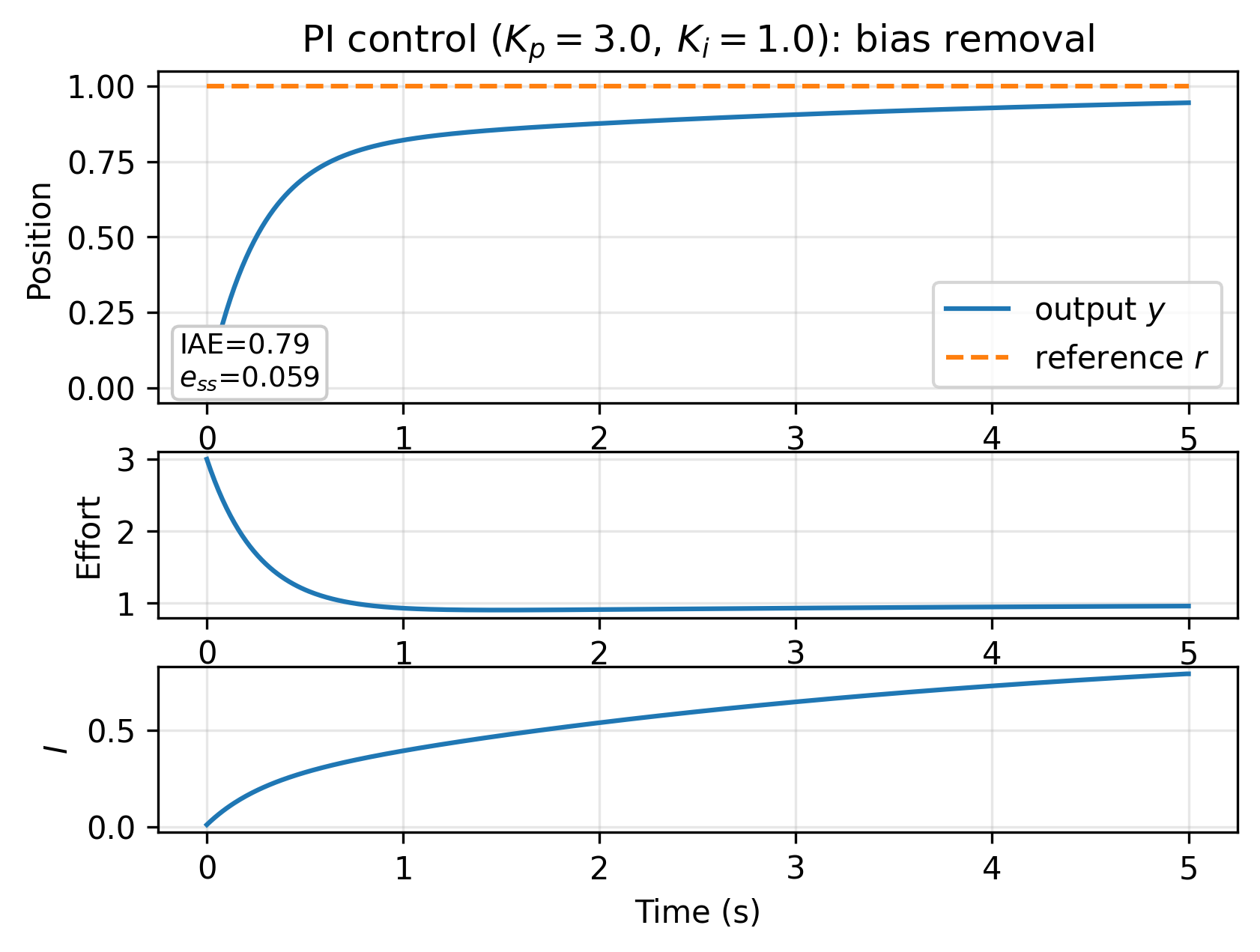}
\caption{PI at $K_p=3.0$, $K_i=1.0$.}
\label{fig:pi_ki1}
\end{figure}

\begin{figure}[!t]
\centering
\includegraphics[width=\columnwidth]{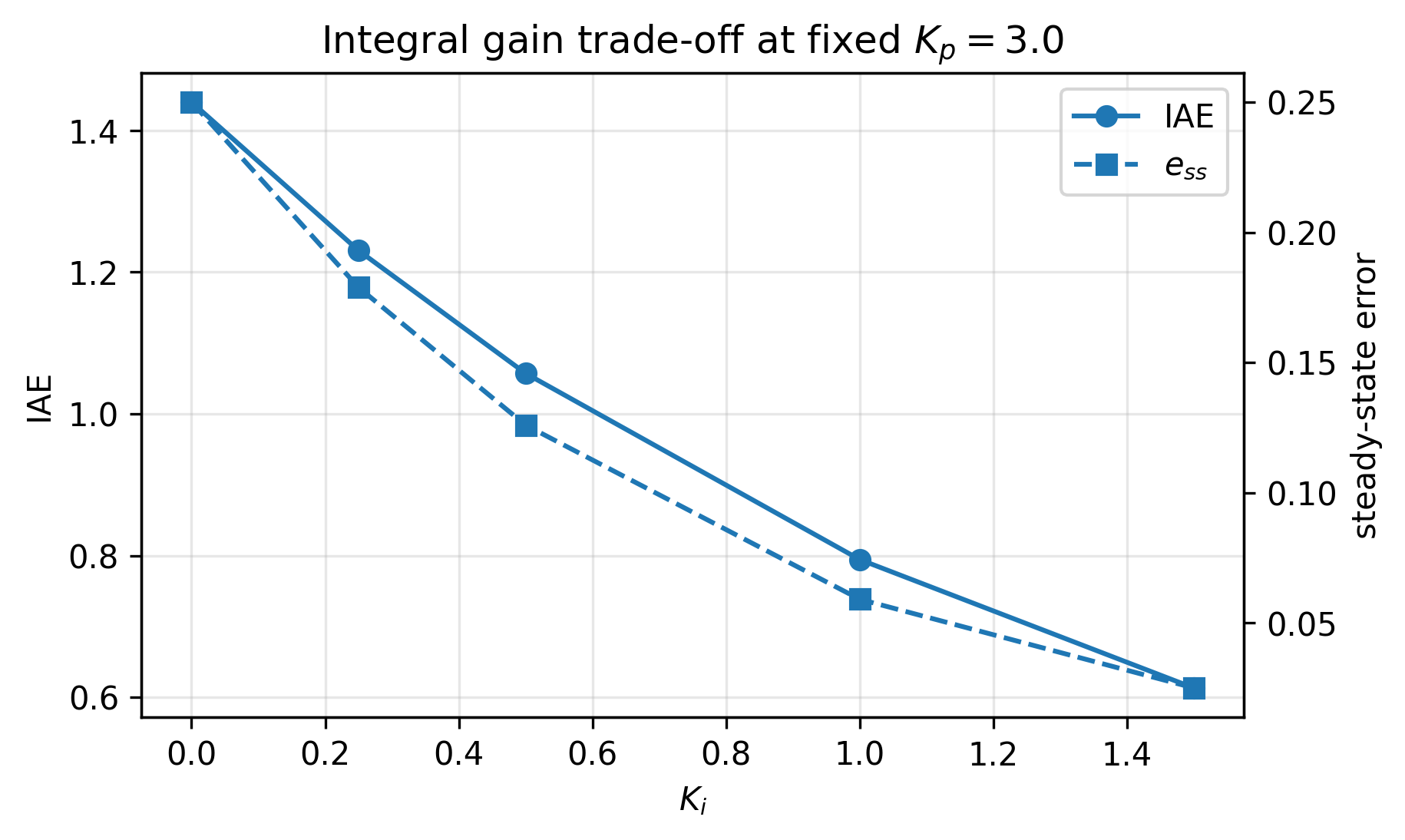}
\caption{IAE vs. $K_i$ at $K_p=3.0$.}
\label{fig:iae_vs_ki}
\end{figure}

Fig.~\ref{fig:pi_baseline}--Fig.~\ref{fig:iae_vs_ki} visualize the effect of increasing $K_i$ at fixed $K_p=3.0$.

\subsubsection{Full PID control}

\begin{table}[!t]
\caption{PID step-response metrics at $K_p=3.0$, $K_i=1.0$.}
\label{tab:pid}
\centering
\resizebox{\columnwidth}{!}{%
\begin{tabular}{@{}rrrrrr@{}}
\toprule
$K_d$ & \%OS & $t_r$ (s) & $t_s$ (s) & $e_{ss}$ & IAE \\
\midrule
0.00 & 0.0000 & 2.7800 & -- & 0.0590 & 0.7789 \\
0.05 & 0.0000 & 2.7300 & -- & 0.0580 & 0.7831 \\
0.10 & 0.0000 & 2.6800 & -- & 0.0570 & 0.7873 \\
\bottomrule
\end{tabular}}
\end{table}

\begin{figure}[!t]
\centering
\includegraphics[width=\columnwidth]{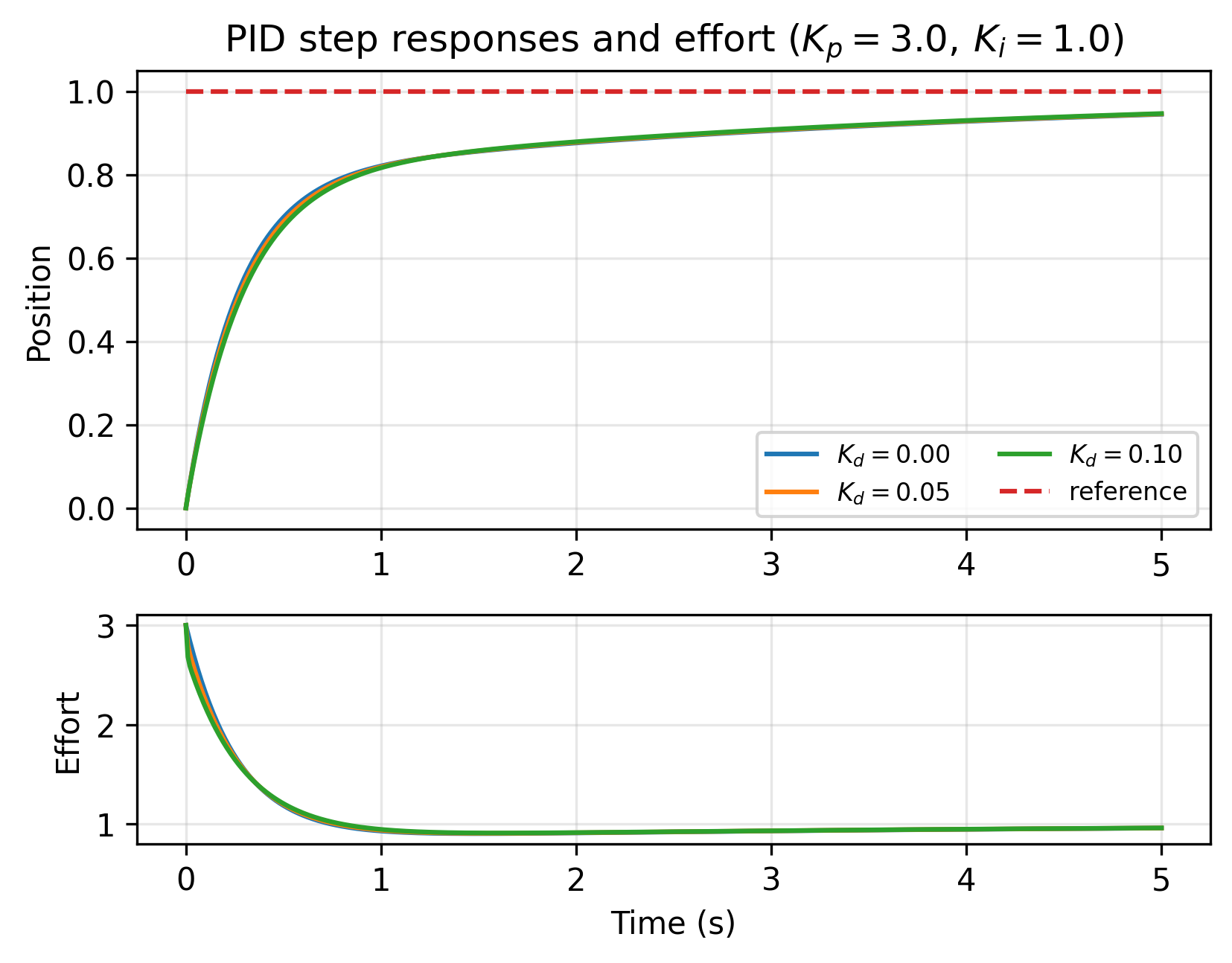}
\caption{PID step responses at $K_p=3.0$, $K_i=1.0$, varying $K_d$ (merged).}
\label{fig:pid_merged}
\end{figure}

\begin{figure}[!t]
\centering
\includegraphics[width=\columnwidth]{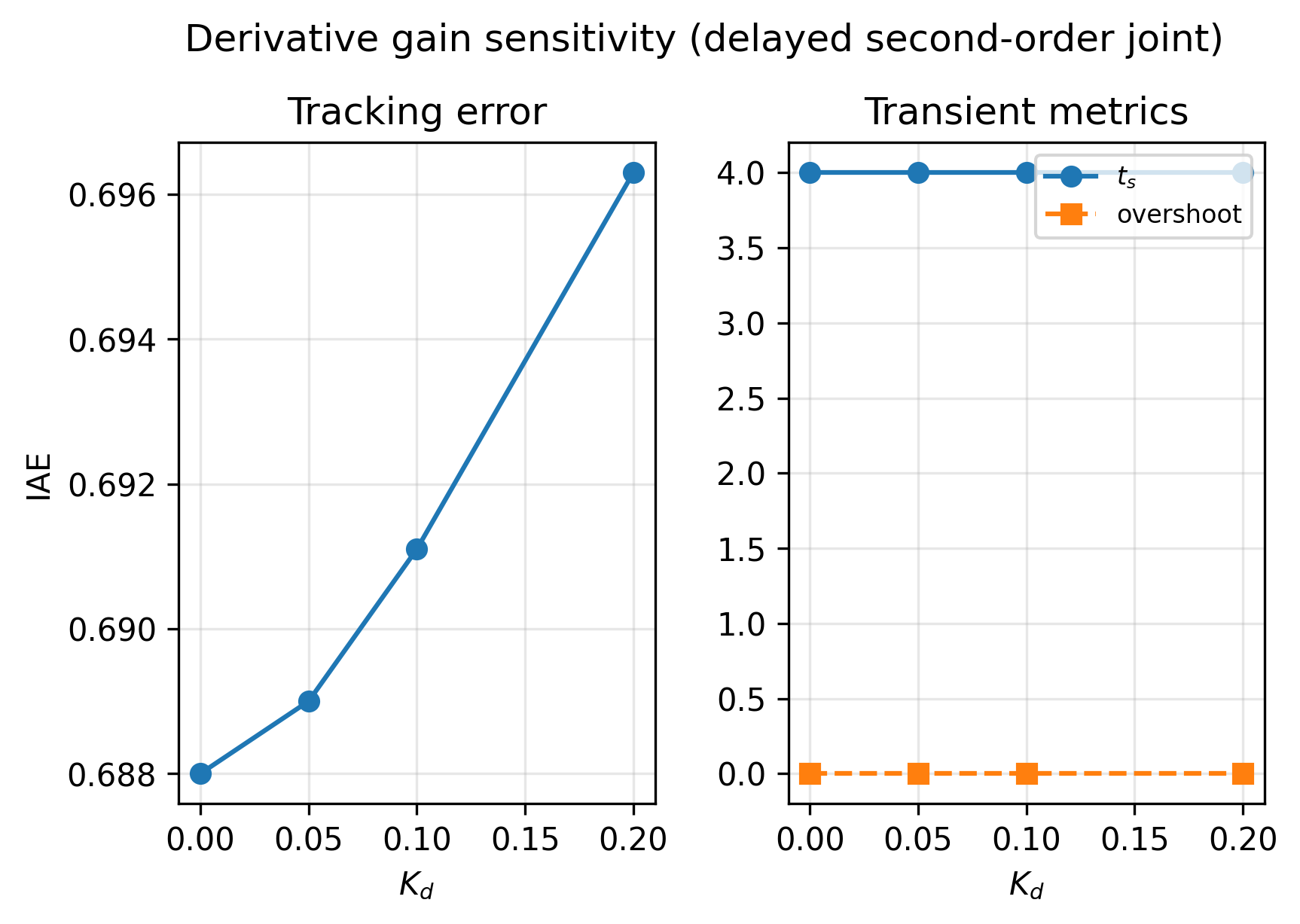}
\caption{Derivative gain sensitivity under a delayed second-order joint model: IAE (left) and transient metrics (right) vs. $K_d$.}
\label{fig:settle_vs_kd}
\end{figure}

Figs.~\ref{fig:pid_merged}--\ref{fig:settle_vs_kd} separate two regimes. On a delay-free single-pole plant, changing $K_d$ has only a second-order influence relative to $K_p$ and $K_i$ (Fig.~\ref{fig:pid_merged}). Once higher-order actuator dynamics and a few samples of delay are introduced, derivative action becomes measurably useful for damping and transient shaping (Fig.~\ref{fig:settle_vs_kd}). For readability, the three step-response traces are merged into a single figure.

\subsection{Saturation-dominant results and anti-windup}
\label{sec:aw_results}
\begin{figure}[!t]
\centering
\includegraphics[width=\columnwidth]{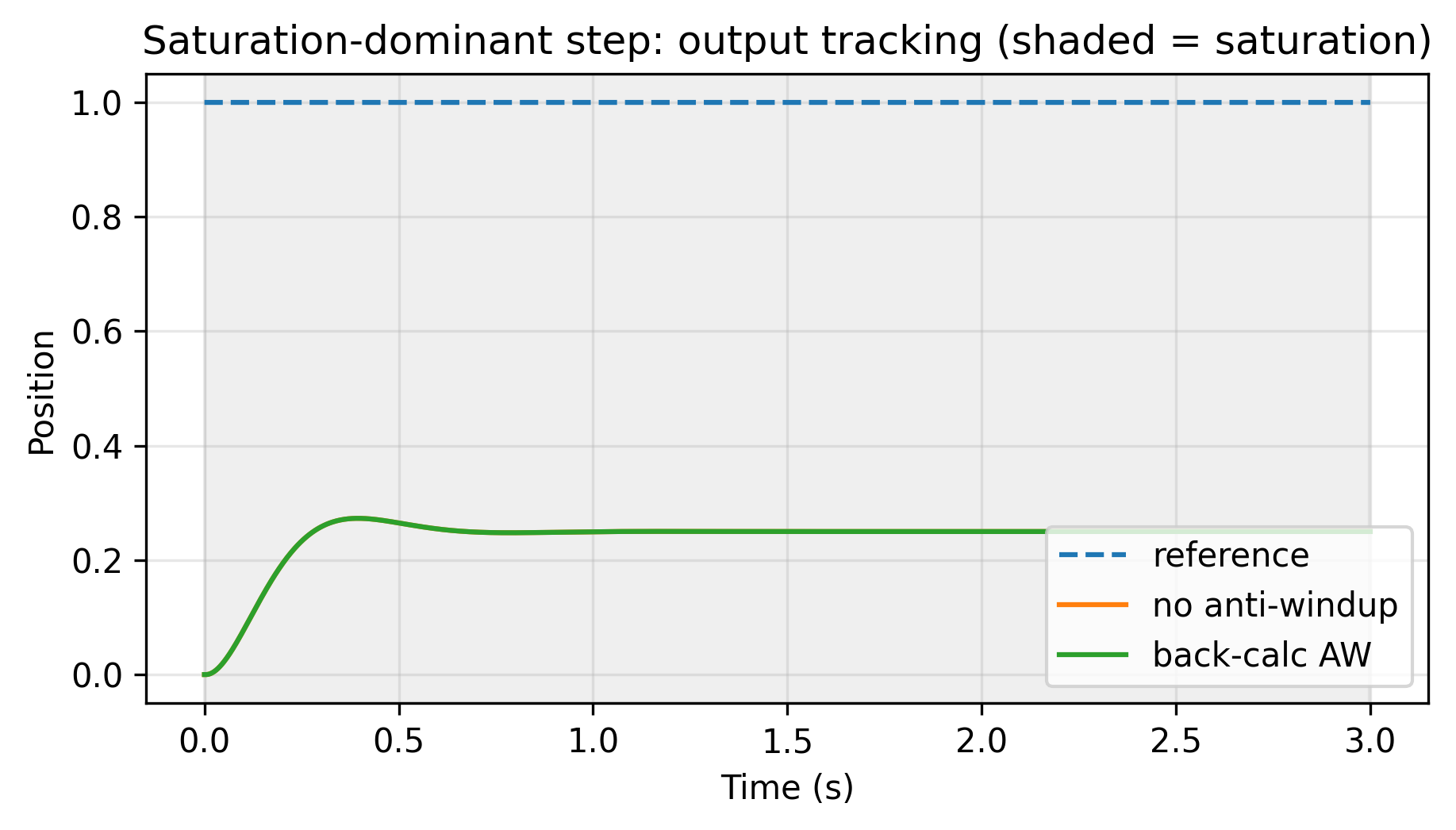}
\caption{Saturation-dominant (output).}
\label{fig:windup_y}
\end{figure}

\begin{figure}[!t]
\centering
\includegraphics[width=\columnwidth]{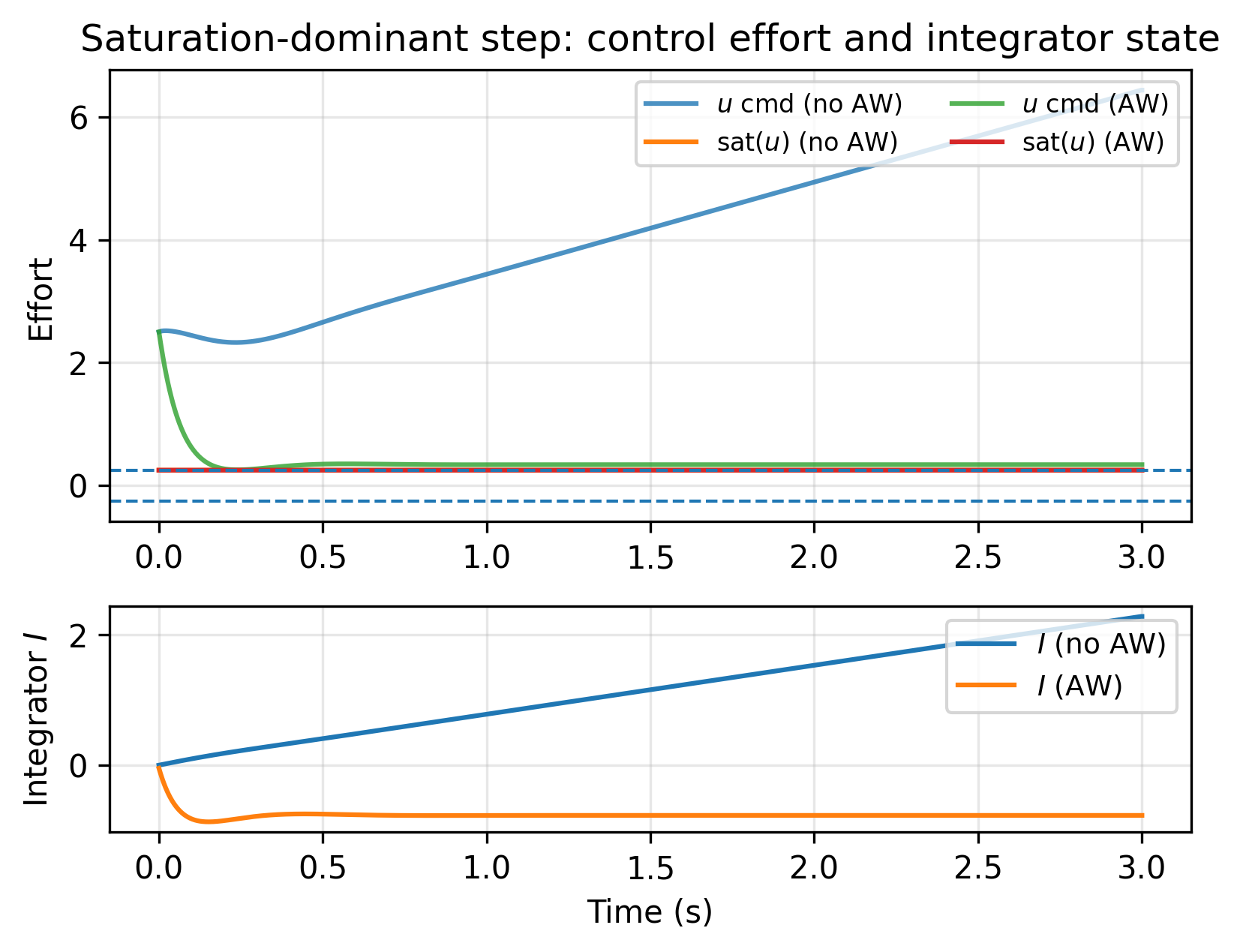}
\caption{Saturation-dominant (effort).}
\label{fig:windup_u}
\end{figure}

Figs.~\ref{fig:windup_y}--\ref{fig:windup_u} report the saturation-dominant scenario described in Sec.~\ref{sec:aw_design}. This test forces the actuator to clamp during the early transient so that, without compensation, the integrator would otherwise accumulate error while the input is pinned at its limit.

\subsection{Monte Carlo robustness results}
\label{sec:mc_results}
We next evaluate the same controller family under the randomized model distribution described in Sec.~\ref{sec:robust_design}. This test probes whether a gain choice that performs well on the nominal model continues to behave acceptably under uncertainty in $(\tau,K)$, sample delay, measurement noise/quantization, and tighter saturation limits.

Table~\ref{tab:mc_summary} summarizes the median performance across the model family for a manual baseline and a robustness-oriented tuned controller.

\begin{table}[!t]
\caption{Monte Carlo robustness summary over a family of joint models (ZOH discretization, randomized $\tau$, $K$, delay, noise, quantization, and saturation limits).}
\label{tab:mc_summary}
\centering
\footnotesize
\resizebox{\columnwidth}{!}{%
\begin{tabular}{@{}lllllll@{}}
\toprule
\textbf{Controller} & \textbf{Kp} & \textbf{Ki} & \textbf{Kd} & \textbf{Median IAE} & \textbf{Median OS (\%)} & \textbf{Median sat duty} \\
\midrule
Manual & 3.000 & 1.000 & 0.050 & 0.687 & 0.000 & 0.000 \\
Robust-tuned & 10.000 & 25.000 & 0.800 & 0.470 & 1.000 & 0.000 \\
\bottomrule
\end{tabular}}
\end{table}

\begin{figure}[!t]
\centering
\includegraphics[width=\columnwidth]{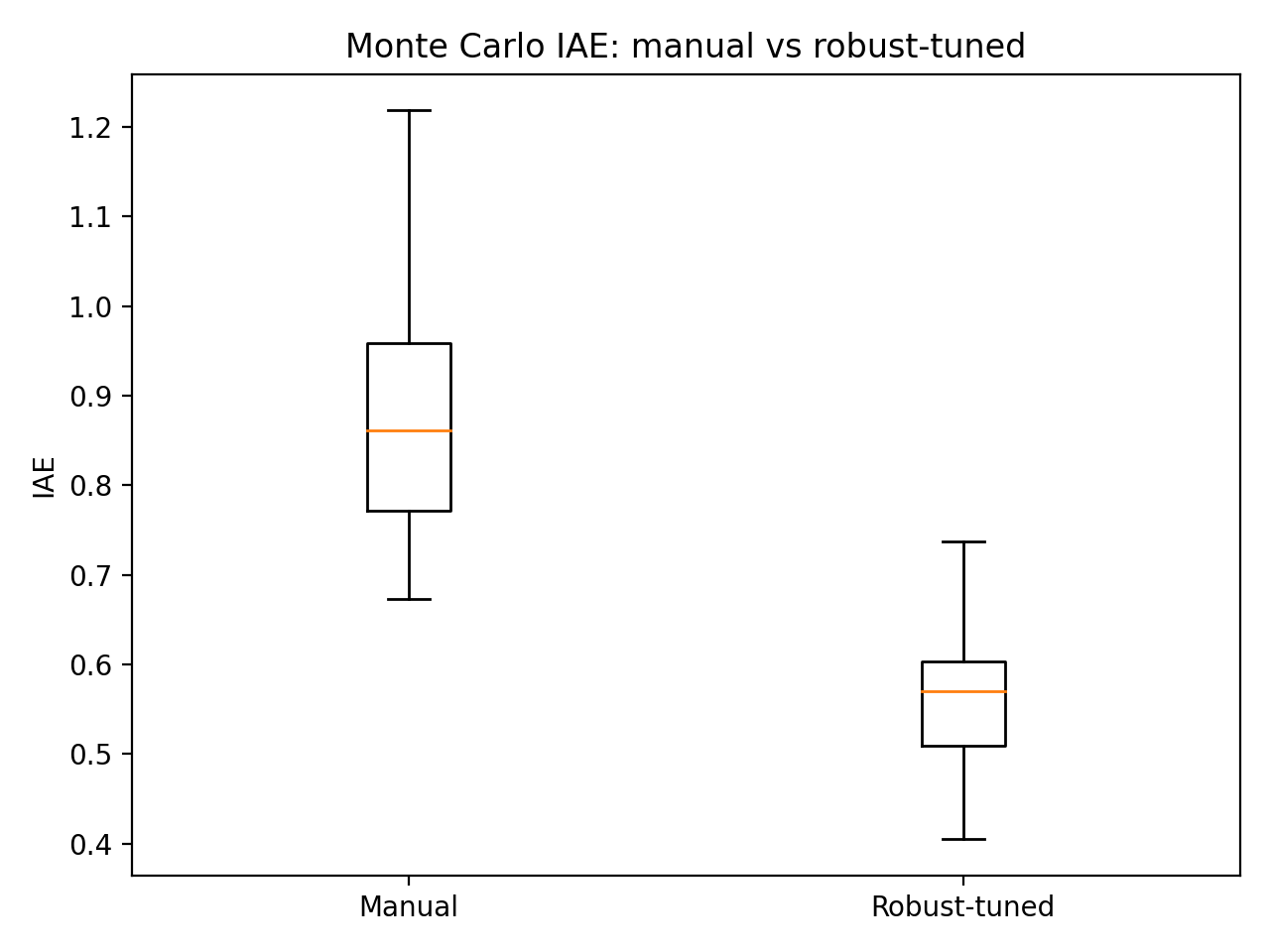}
\caption{Monte Carlo IAE: manual vs robust-tuned.}
\label{fig:mc_iae}
\end{figure}

\begin{figure}[!t]
\centering
\includegraphics[width=\columnwidth]{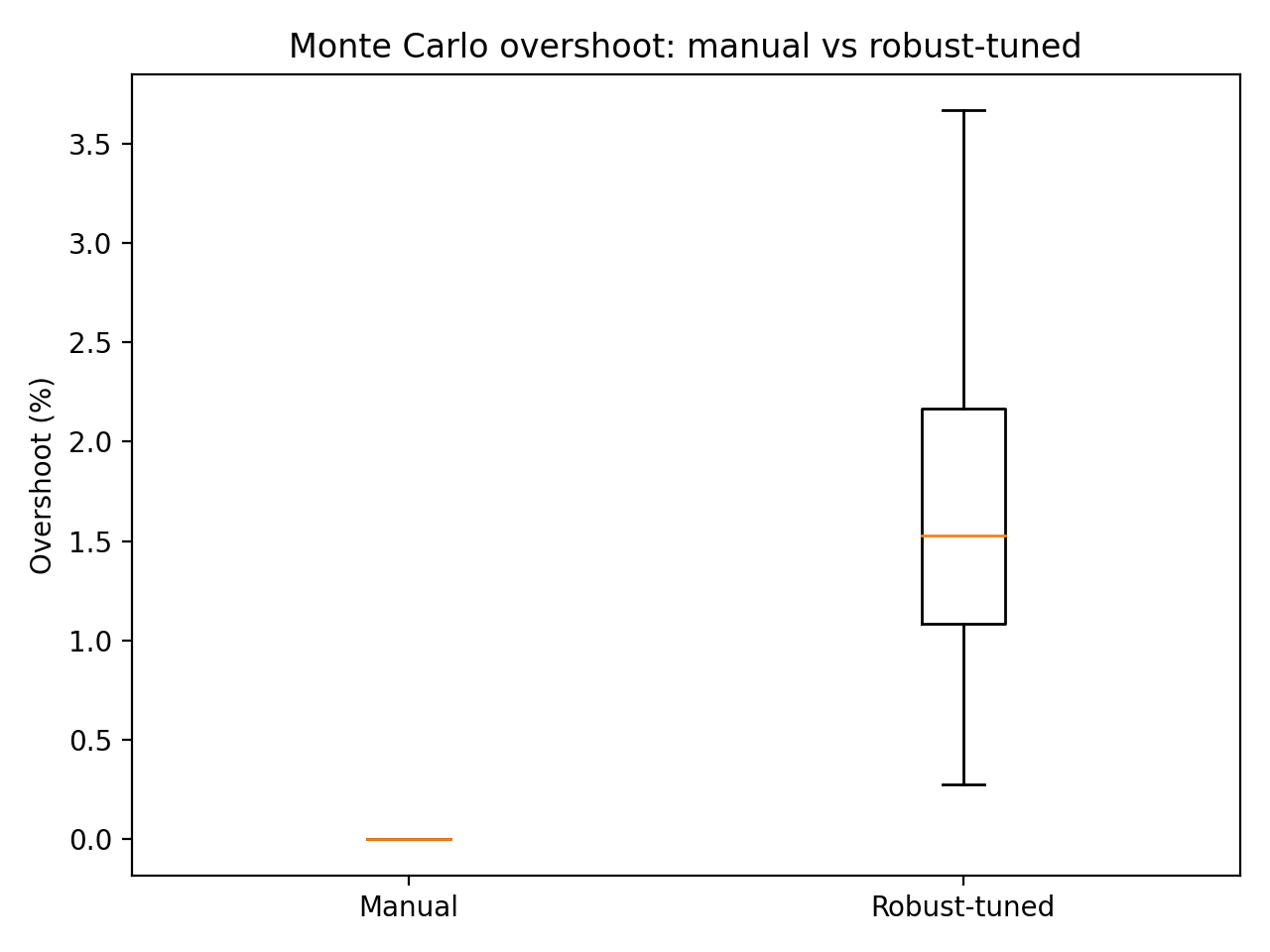}
\caption{Monte Carlo overshoot: manual vs robust-tuned.}
\label{fig:mc_os}
\end{figure}

\begin{figure}[!t]
\centering
\includegraphics[width=\columnwidth]{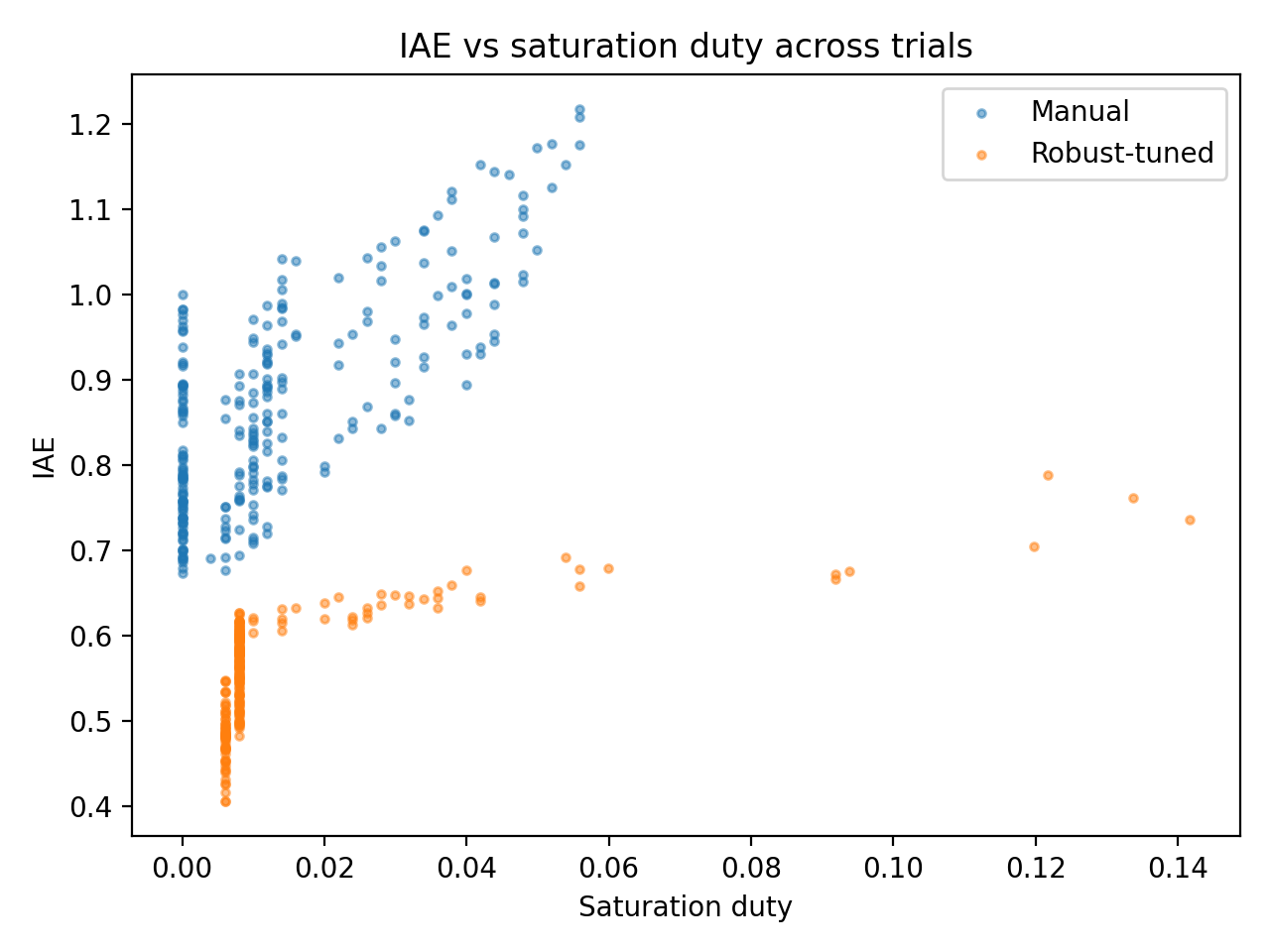}
\caption{IAE vs saturation duty across trials.}
\label{fig:pareto_scatter}
\end{figure}

\subsection{Expanded-benchmark Safe-BO results}
\label{sec:bo_results}
We now report results for the hybrid-certified Safe-BO workflow described in Sec.~\ref{sec:bo_workflow} when evaluated on the expanded benchmark suite (first-order ZOH family + second-order actuator family). This experiment addresses two practical questions: (i) does the certification filter prevent clearly unacceptable evaluations while still allowing progress, and (ii) does the constrained workflow remain sample-efficient compared to unconstrained search?

\begin{figure}[!t]
\centering
\includegraphics[width=\columnwidth]{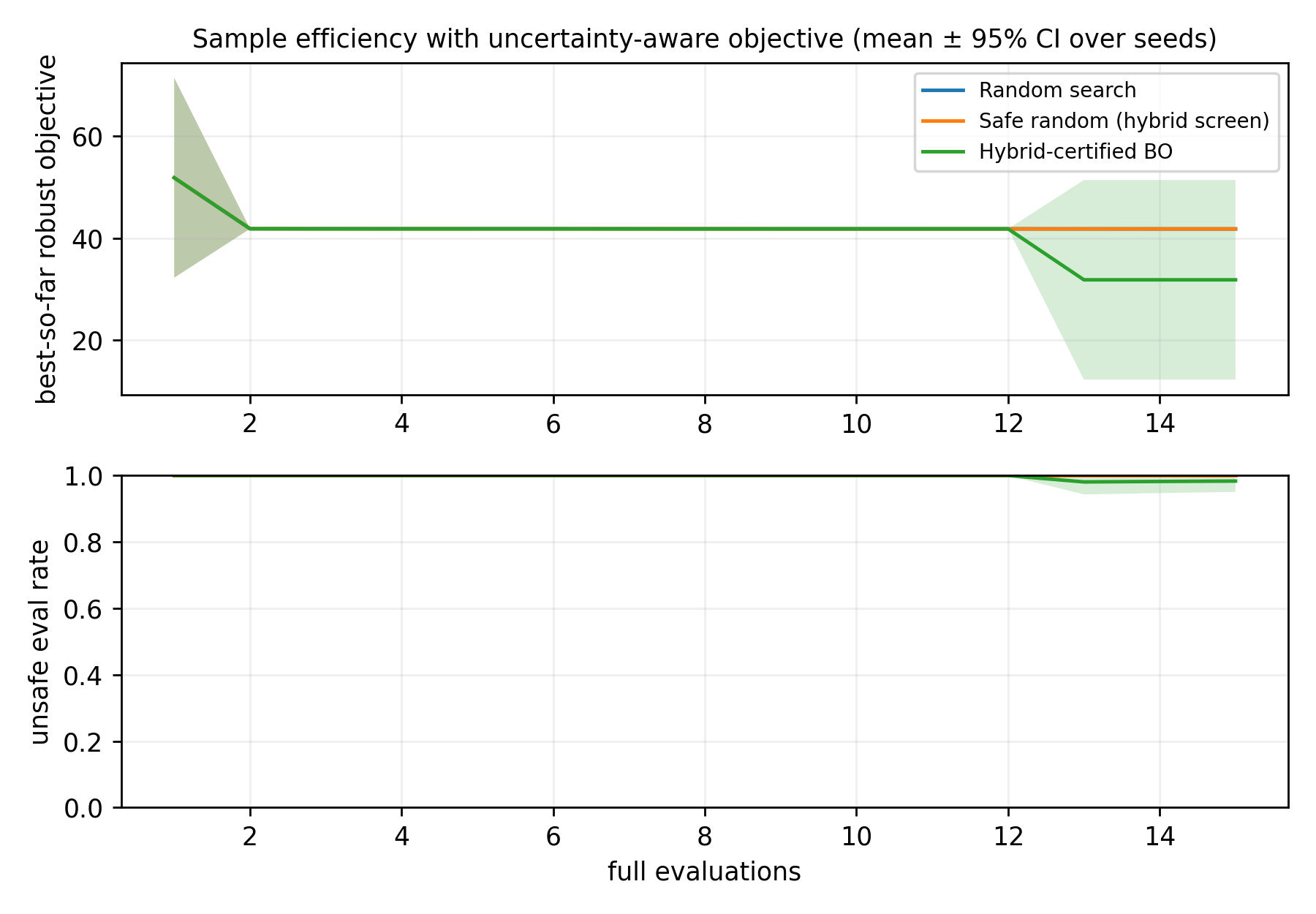}
\caption{Expanded benchmark: best-so-far robust objective vs. evaluations (mean $\pm$ 95\% CI over seeds, top) and unsafe evaluation rate (bottom).}
\label{fig:exp_bo_curve}
\end{figure}

\begin{figure}[!t]
\centering
\includegraphics[width=\columnwidth]{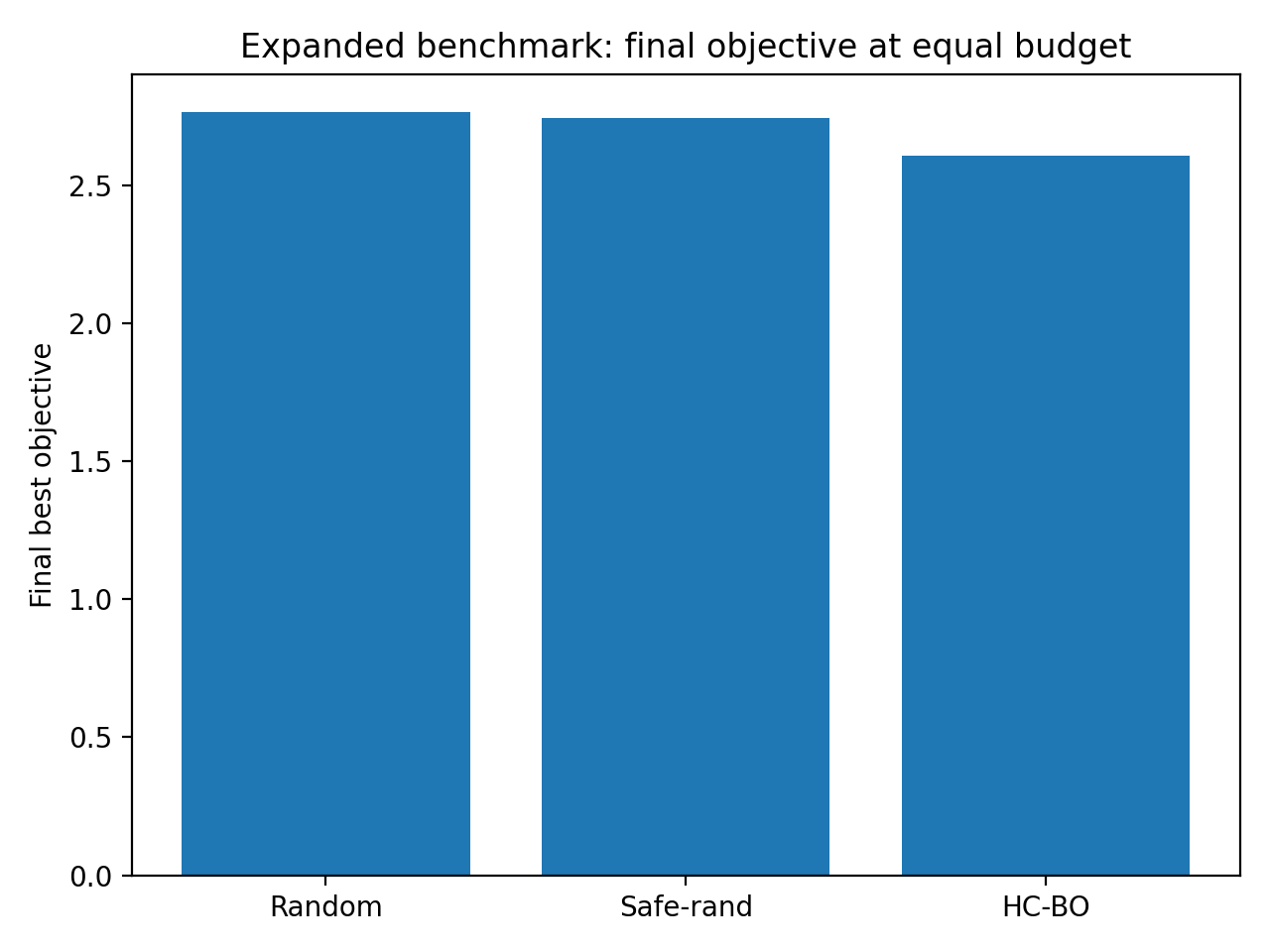}
\caption{Expanded benchmark: final objective at equal budget.}
\label{fig:exp_bo_bar}
\end{figure}

\begin{table}[t]
\caption{Representative second-order actuator family used for the benchmark plots (Figs.~\ref{fig:benchmark_step}--\ref{fig:benchmark_aw}).}
\label{tab:second_order_params}
\centering
\begin{tabular}{@{}ll@{}}
\toprule
Parameter & Value (nominal) \\
\midrule
Natural frequency $\omega_n$ & $8$--$10$ rad/s \\
Damping ratio $\zeta$ & $0.6$--$0.8$ \\
Input gain $K_u$ & $1.0$ \\
Sample period $\Delta t$ & $2$ ms \\
Input saturation $u_{\max}$ & $1.0$ (step/sine), $0.25$ (AW ablation) \\
Sample delay & $1$ sample \\
Friction & viscous $b=0.05$--$0.06$, Coulomb $f_c=0.02$--$0.03$ \\
Reference & unit step or $0.5\sin(2\pi\cdot0.8t)$ \\
\bottomrule
\end{tabular}
\end{table}

\begin{figure}[!t]
\centering
\includegraphics[width=\columnwidth]{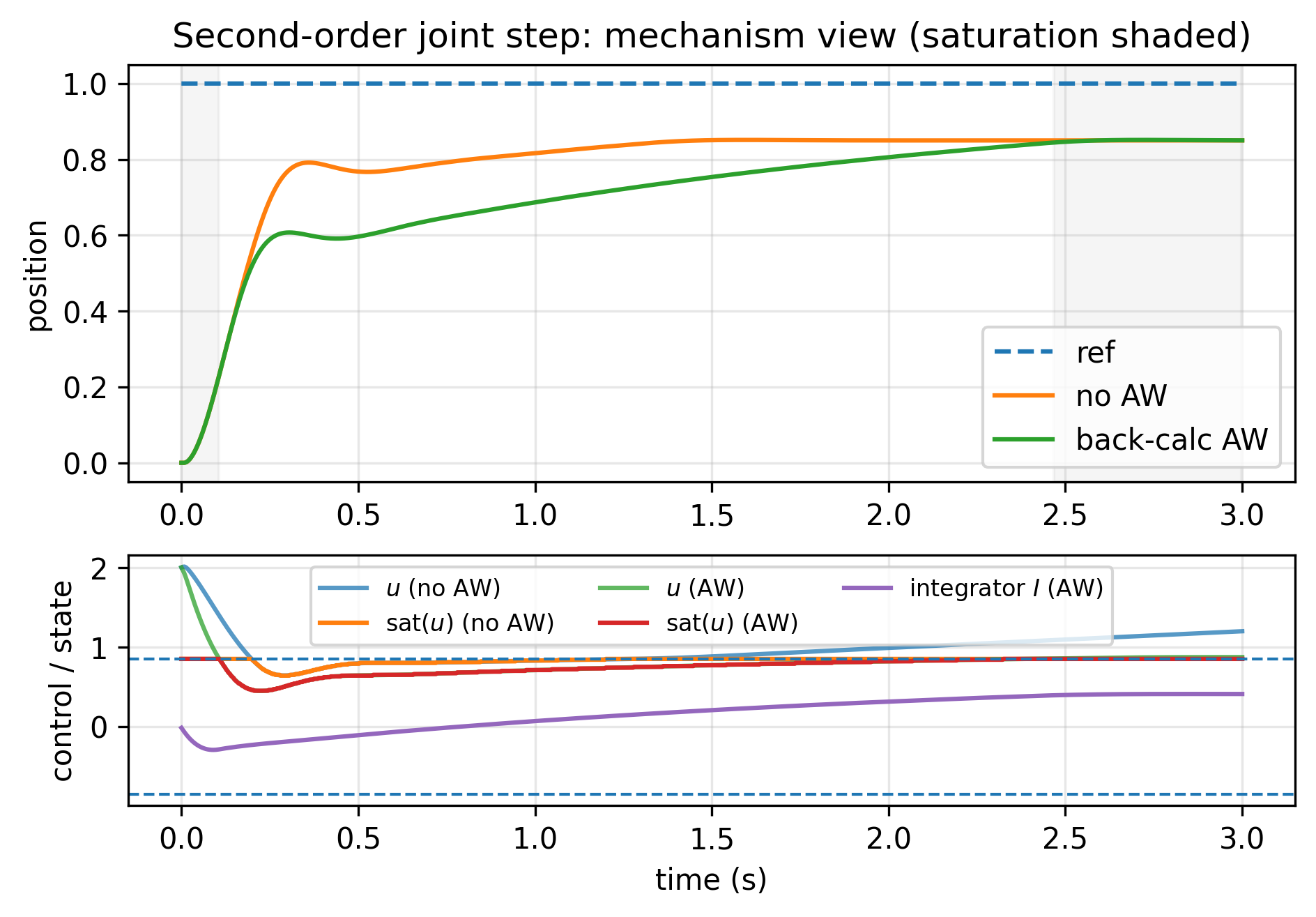}
\caption{Second-order actuator benchmark (step): output tracking with saturation intervals shaded (top) and corresponding commanded/saturated control plus integrator state (bottom), comparing no anti-windup vs. back-calculation.}
\label{fig:benchmark_step}
\end{figure}

\begin{figure}[!t]
\centering
\includegraphics[width=\columnwidth]{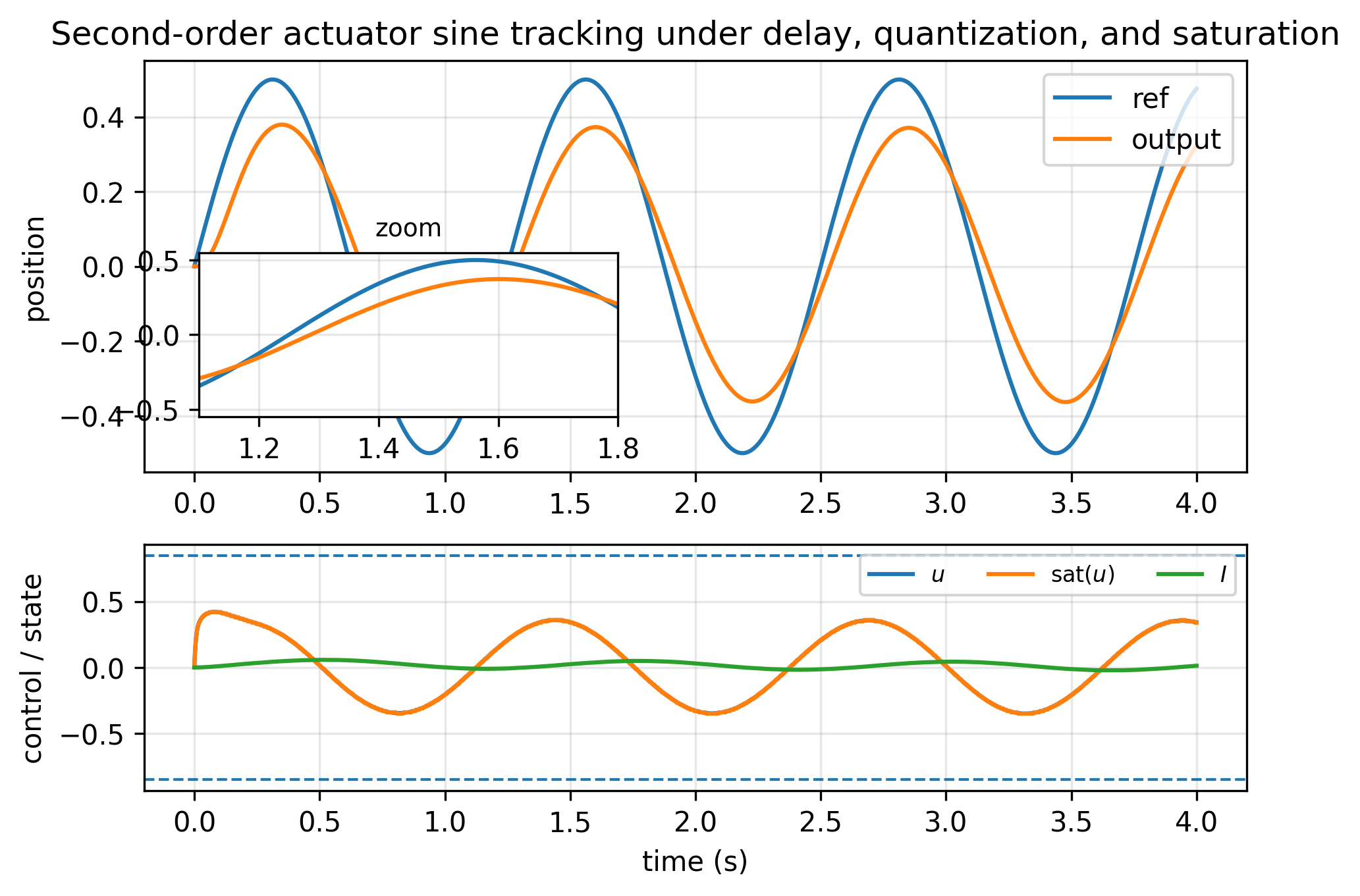}
\caption{Second-order actuator benchmark (sine): tracking under delay, quantization, and saturation (top, with inset zoom) and corresponding commanded/saturated control plus integrator state (bottom).}
\label{fig:benchmark_sine}
\end{figure}

\begin{figure}[!t]
\centering
\includegraphics[width=\columnwidth]{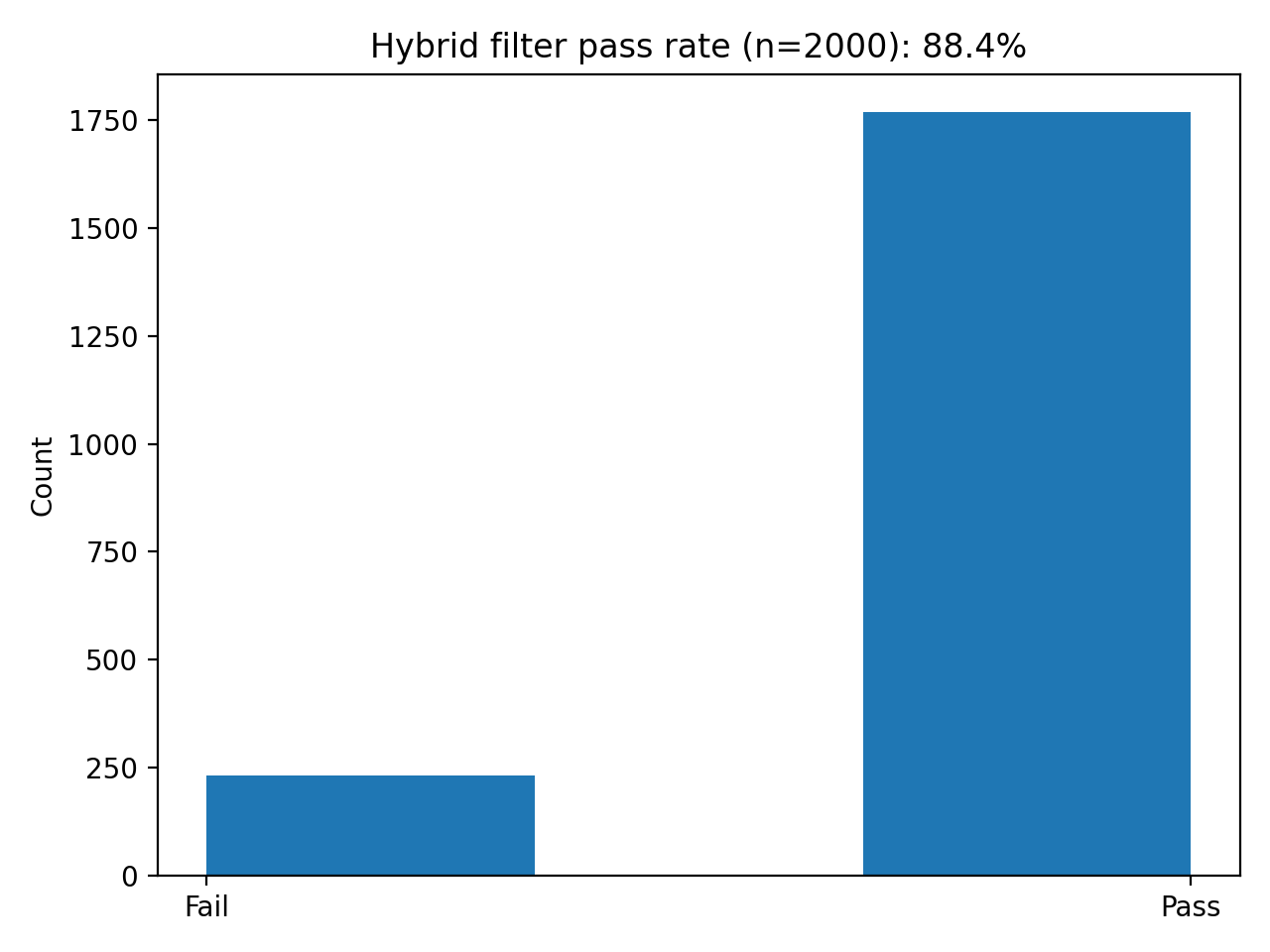}
\caption{Feasible fraction passing hybrid filter.}
\label{fig:feasible_fraction}
\end{figure}

\begin{figure}[!t]
\centering
\includegraphics[width=\columnwidth]{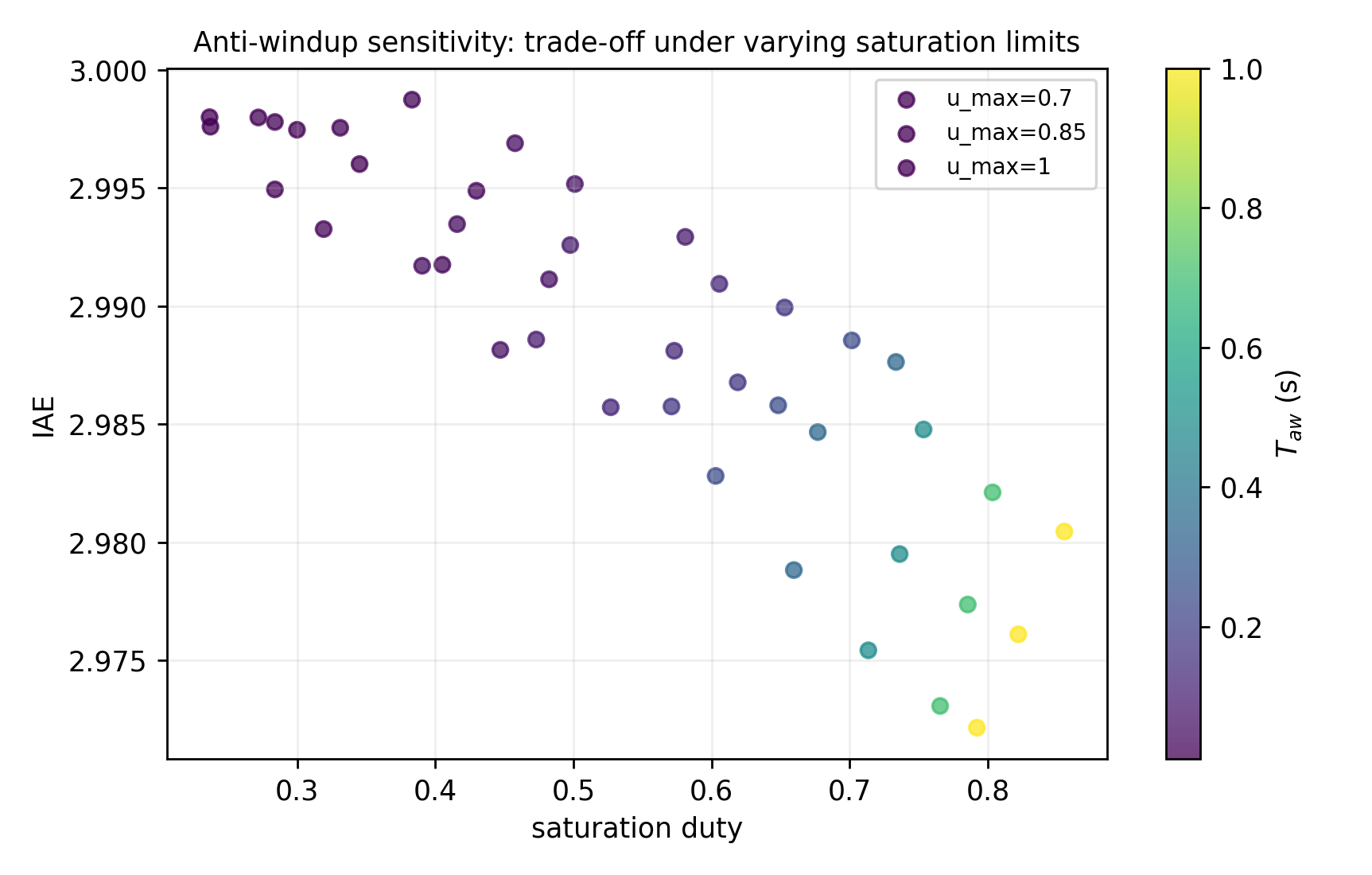}
\caption{Anti-windup sensitivity as a trade-off between IAE and saturation duty across back-calculation time constants $T_{aw}$ (color) and multiple saturation limits $u_{\max}$.}
\label{fig:benchmark_aw}
\end{figure}

\section{Discussion}
Baseline sweeps confirm that for a delay-free first-order plant, $K_p$ primarily sets the steady-state tail (but cannot remove steady-state error), and $K_i$ dominates tail removal and IAE reduction. On this nominal single-pole system, derivative action yields only modest shaping; however, once delay and higher-order actuator dynamics are introduced, $K_d$ becomes materially beneficial for damping and transient control. The stability-region derivations provide a fast guardrail for PI gains before evaluation, and the saturation-dominant tests demonstrate why anti-windup is essential once the clamp is active. Robustness-oriented tuning and Safe-BO results illustrate a practical workflow for constrained tuning: optimizing gains against randomized model families improves typical-case tracking while keeping overshoot bounded and rejecting a large fraction of unsafe candidate gains.

\subsection{Limitations and threats to validity}
The results emphasize that sampling, saturation, and small delays change both the safe gain set and the qualitative closed-loop behavior, even for simple plants. The ZOH and Euler PI regions provide an interpretable first-line guardrail for the integral action that most strongly governs steady-state and tail behavior.

\noindent\textbf{Scope.} The analytic guard is nominal and PI-specific; for full PID tuning we therefore rely on the behavioral screen and robust evaluation, and we report evaluation-based evidence rather than hardware guarantees. We explicitly model uncertainty, delay, noise, and quantization to approximate common embedded joint loops, but the benchmark models remain simplified abstractions.

\section{Conclusion}
This paper reformulates a classic PID tuning problem in the practical setting of discrete-time embedded control with actuator limits. In addition to baseline P/PI/PID characterization, we derive PI stability regions using the Jury criterion, demonstrate saturation-dominant anti-windup behavior, and propose a robustness-oriented tuning workflow that emulates key non-idealities through a randomized model family. We further introduce a hybrid-certified Safe-BO workflow to prevent unsafe gain exploration and demonstrate its behavior on a second-order actuator benchmark suite.

\appendix
\section{Exact zero-order-hold discretization}
Forward Euler is simple and matches many educational implementations, but a first-order LTI plant also admits an exact discrete-time update under a ZOH assumption:
\begin{equation}
y_{k+1} = a\,y_k + b\,u_k,\quad a=e^{-\Delta t/\tau},\quad b=K\left(1-e^{-\Delta t/\tau}\right).
\label{eq:zoh}
\end{equation}

\section{Implementation notes for embedded robotic joints}
This section summarizes practical details that materially affect digital PID behavior but are often omitted in simplified analyses.

\subsection{Derivative filtering and noise}
The derivative term amplifies measurement noise. In embedded implementations, a first-order low-pass filter (``dirty derivative'') is commonly used:
\begin{equation}
D(s)=\frac{K_d s}{1+\frac{s}{\omega_f}} ,
\end{equation}
implemented in discrete time using a stable recursion. In our evaluation studies, adding measurement noise shifts the trade-off between overshoot and IAE, and derivative action becomes beneficial only when the effective bandwidth is chosen relative to the sampling period and noise level. As a practical guideline, we recommend selecting $\omega_f$ such that $\omega_f T_s \in [0.1,\,0.3]$ and evaluating robustness under noise/quantization before increasing $K_d$.

\subsection{Anti-windup tuning and saturation duty}
Back-calculation anti-windup introduces an additional time constant $T_{aw}$ (or gain $K_{aw}$). Setting $T_{aw}$ too small can reintroduce aggressive behavior near saturation; setting it too large yields slow recovery tails. We therefore report a saturation-duty metric (fraction of time in saturation) alongside IAE and overshoot, and we treat ``low duty + low IAE'' as a robustness indicator under tighter effort limits.

\subsection{Delay, quantization, and discrete-time guards}
Computational delay and quantization act like additional dynamics that reduce phase margin. We therefore propose using analytic discrete-time stability guards (Jury conditions) as a prefilter for any automated tuning loop, and then validating candidates under randomized uncertainty (delay, quantization, noise) before declaring a gain set robust.

\subsection{Reproducibility checklist}
\begin{itemize}
\item Plant model family and uncertainty bounds: $(K,\tau)$ ranges; delay/jitter distribution; noise and quantization levels.
\item Digital implementation: sampling period $T_s$; discretization method (Euler/ZOH); derivative filter ($\omega_f$); anti-windup form and parameter ($T_{aw}$ or $K_{aw}$).
\item Actuator constraints: saturation bounds $u_{\max}$ (and any rate limits if used); definition of saturation duty.
\item Task definition: reference trajectory, horizon length, and any disturbances/payload changes.
\item Metrics: IAE definition and normalization; overshoot threshold; penalties/weights; early-termination rules for unsafe trials.
\item Optimization protocol: candidate bounds, initial design size, BO kernel/acquisition, and random seeds.
\end{itemize}

\subsection{Objective design for tuning}
Robotics PID tuning is frequently framed as ``make tracking good,'' but the objective choice strongly influences the resulting gains. A pure tracking-error objective tends to push gains toward aggressive behavior that exploits saturation, whereas adding explicit penalties on effort and effort rate discourages chattering and reduces thermal stress. In this work, we evaluate candidates under a composite objective that includes IAE, overshoot, and a saturation-duty term; this combination yields a predictable trade-off between tracking performance and saturation exposure.

\subsection{Reporting standards for reproducible tuning}
For papers that propose tuning procedures (manual or automated), we recommend reporting: the target trajectory, the error metric (and whether it is absolute/squared), the evaluation horizon, saturation bounds, sampling period, delay model, and sensor quantization/noise assumptions. Without these details, it is difficult to interpret whether a reported ``better'' controller is due to a tuning method or due to a different experimental setup.

\section{Jury-test coefficients for PI with ZOH}
For completeness, we outline the coefficient mapping used for the ZOH-discretized first-order plant. Let $G(s)=\frac{K}{\tau s+1}$ and let the ZOH equivalent at sampling period $T_s$ be
\begin{equation}
G(z)=\frac{K\,(1-e^{-T_s/\tau})}{z-e^{-T_s/\tau}} .
\end{equation}
With the discrete PI controller $C(z)=K_p + K_i \frac{T_s z}{z-1}$, the closed-loop characteristic polynomial can be written in the monic form
\begin{equation}
p(z)=z^2 + a_1 z + a_0 ,
\end{equation}
where $a_1$ and $a_0$ are affine functions of $(K_p,K_i)$ for fixed $(K,\tau,T_s)$. The Jury stability test for a second-order polynomial reduces to three inequalities:
\begin{equation}
1+a_1+a_0>0,\qquad 1-a_1+a_0>0,\qquad 1-a_0>0.
\end{equation}
These constraints define a convex polygon in the $(K_p,K_i)$ plane and serve as an analytic prefilter in our hybrid-certified search.

\section{Hybrid-certified Bayesian optimization procedure}
We summarize the hybrid-certified tuning workflow used in our experiments.
\begin{enumerate}
\item \textbf{Define} a feasible gain box for $(K_p,K_i,K_d)$ and implementation parameters (sampling period $T_s$, derivative filter bandwidth $\omega_f$, and anti-windup time constant $T_{aw}$).
\item \textbf{Analytic prefilter:} reject any candidate that violates the discrete-time stability guards derived for the nominal model (Jury conditions for PI, and corresponding linearized constraints for PID).
\item \textbf{Behavioral check:} run a short-horizon evaluation on a small subset of uncertain models. Reject candidates that produce divergence, excessive overshoot, or persistent saturation.
\item \textbf{Robust scoring:} evaluate remaining candidates on a larger uncertainty set and compute a robust objective (median IAE plus penalties in this paper; a risk-averse CVaR variant can be substituted when tail performance dominates).
\item \textbf{BO update:} update the Gaussian-process surrogate and propose the next candidate, repeating until the evaluation budget is exhausted.
\end{enumerate}
This structure preserves BO's data efficiency while enforcing stability and practicality constraints that are critical for robotic joints.

\FloatBarrier

\end{document}